%% file: main.tex
\pdfoutput=1
\documentclass[10pt,twocolumn,letterpaper]{article}

\usepackage{cvpr}               % To produce the CAMERA-READY version
\usepackage[accsupp]{axessibility}

\input{preamble}

\def\cvprPaperID{3980} % *** Enter the CVPR Paper ID here
\def\confName{CVPR}
\def\confYear{2023}

\begin{document}

\title{Habitat-Matterport 3D Semantics Dataset}

\author{
\textbf{Karmesh Yadav$^{1}$\thanks{Equal Contribution, Correspondence: \url{ykarmesh@gmail.com}} , Ram Ramrakhya$^{2}$\footnotemark[1] , Santhosh Kumar Ramakrishnan$^{3}$\footnotemark[1] ,}\\
\textbf{Theo Gervet$^{6}$, John Turner$^{1}$, Aaron Gokaslan$^{4}$, Noah Maestre$^{1}$, Angel Xuan Chang$^{5}$, }\\
\textbf{Dhruv Batra$^{1,2}$, Manolis Savva$^{5}$, Alexander William Clegg$^{1}$\thanks{Equal Contribution} , Devendra Singh Chaplot$^{1}$\footnotemark[2]}\\
$^{1}$Meta AI~~$^{2}$Georgia Tech~~$^{3}$UT Austin\\
$^{4}$Cornell University~~$^{5}$Simon Fraser University~~$^{6}$Carnegie Mellon University
}

\maketitle

\input{fig/teaser}

\input{sections/abstract}
\input{sections/introduction}
\input{sections/related}
\input{sections/dataset}
\input{sections/experiments}
\input{sections/challenge}

\input{sections/conclusion}
\input{sections/acknowledgments}

\clearpage

{\small
\bibliographystyle{unsrtnat}
\setlength{\bibsep}{0pt}
\bibliography{main}
}

\clearpage
\newpage
\appendix
\input{sections/appendix}

\end{document}

%% file: preamble.tex
\usepackage[utf8]{inputenc} % allow utf-8 input
\usepackage[T1]{fontenc}    % use 8-bit T1 fonts

\usepackage{times}
\usepackage{epsfig}
\usepackage{graphicx}
\usepackage{amsmath}
\usepackage{amsfonts}
\usepackage{amssymb}
\usepackage{microtype}
\usepackage{appendix}

\usepackage{epigraph}
\usepackage{paralist}
\usepackage{comment}
\usepackage{nicefrac}
\usepackage{tabularx}
\usepackage{booktabs}
\usepackage{array}
\usepackage{float}
\usepackage{multirow}
\usepackage{subcaption}
\usepackage{xspace}
\usepackage{siunitx}
\usepackage{bm} % bold math
\usepackage[table,xcdraw]{xcolor}
\usepackage{capt-of}
\usepackage{caption}
\usepackage{relsize}
\usepackage[numbers,sort,compress]{natbib}
\usepackage{dirtytalk}
\usepackage{csquotes}
\usepackage[export]{adjustbox} % Adjust cell-alignment within table
\usepackage{placeins}
\usepackage{enumitem}
\setlist{nosep}

\usepackage[pagebackref=true,breaklinks=true,colorlinks,bookmarks=false]{hyperref}
\usepackage{url}
\usepackage{cleveref}
\usepackage{xspace}
\usepackage{footnote}
\usepackage{adjustbox}

\crefname{section}{Sec.}{Secs.}
\Crefname{section}{Section}{Sections}
\Crefname{table}{Table}{Tables}
\crefname{table}{Tab.}{Tabs.}

\makeatletter
\DeclareRobustCommand\onedot{\futurelet\@let@token\@onedot}
\def\@onedot{\ifx\@let@token.\else.\null\fi\xspace}

\DeclareMathSymbol{@}{\mathord}{letters}{"3B}

\newcommand{\xhdr}[1]{\vspace{2pt}\noindent\textbf{#1}}
\newcommand{\code}[1]{\texttt{\small #1}}
\mathchardef\mhp="2D

\newcommand{\hmtdsem}{\textsc{HM3DSem}\xspace}

\newcommand{\numscenes}{216\xspace}
\newcommand{\numtrainscenes}{145\xspace}
\newcommand{\numvalscenes}{36\xspace}
\newcommand{\numtestscenes}{35\xspace}

\newcommand{\numrooms}{3@100\xspace}
\newcommand{\numobjects}{142@646\xspace}
\newcommand{\numrawcategories}{1@625}

\newcolumntype{Y}{>{\centering\arraybackslash}X}
\newcolumntype{Z}{>{\centering\arraybackslash\hsize=.5\hsize}X}

\newcounter{magicrownumbers}
\preto\tabular{\setcounter{magicrownumbers}{0}}

\newcolumntype{P}[1]{>{\centering\arraybackslash}p{#1}}
\newcommand{\tb}[1]{\textbf{#1}}

%% file: fig/teaser.tex
\begin{figure*}
\centering
\includegraphics[width=\linewidth]{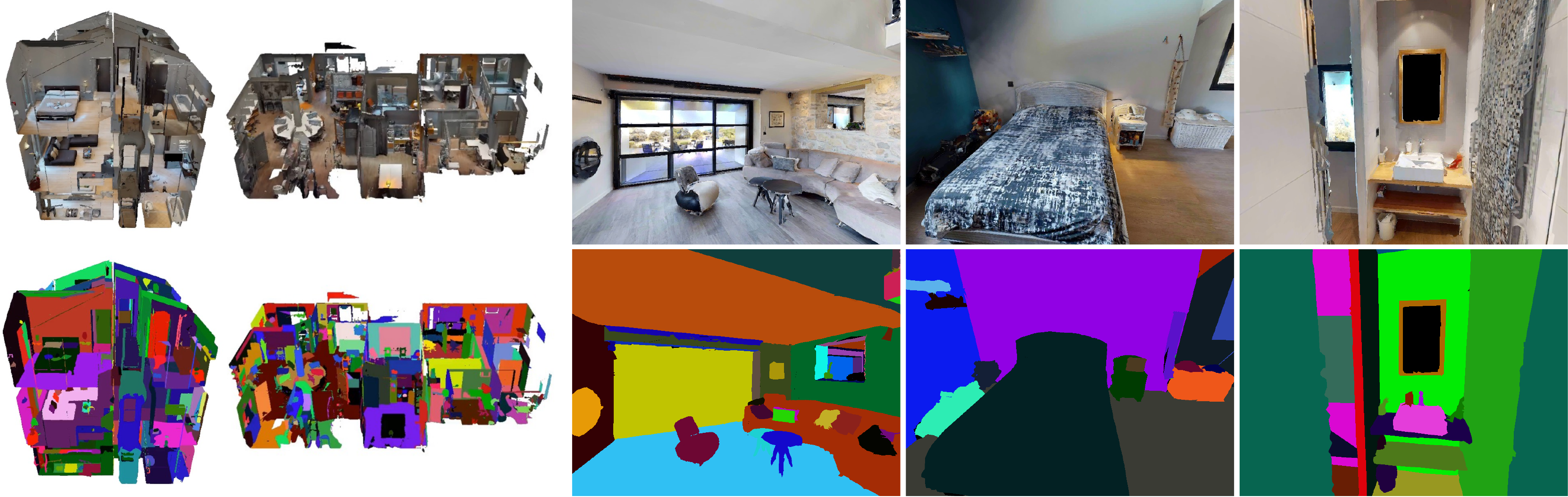}
\caption{\small Habitat-Matterport 3D Semantics (\hmtdsem) provides the largest dataset of real-world spaces with densely annotated semantics.
High-fidelity textured 3D mesh reconstructions are labeled with precise instance-level object semantics, indicated by distinct colors.
}
\label{fig:teaser}
\vspace{-10pt}
\end{figure*}

%% file: sections/abstract.tex
\begin{abstract}
We present the Habitat-Matterport 3D Semantics (\hmtdsem) dataset.
\hmtdsem is the largest dataset of 3D real-world spaces with densely annotated semantics that is currently available to the academic community. It consists of 142,646 object instance annotations across 216 3D spaces and 3,100 rooms within those spaces.
The scale, quality, and diversity of object annotations far exceed those of prior datasets.
A key difference setting apart \hmtdsem from other datasets is the use of texture information to annotate pixel-accurate object boundaries. We demonstrate the effectiveness of \hmtdsem dataset for the Object Goal Navigation task using different methods. Policies trained using \hmtdsem perform outperform those trained on prior datasets. Introduction of \hmtdsem in the Habitat ObjectNav Challenge lead to an increase in participation from 400 submissions in 2021 to 1022 submissions in 2022. Project page: {\small\url{https://aihabitat.org/datasets/hm3d-semantics/}}
% \todo{Expand with summary statistics and findings from experiments.}
\end{abstract}

%% file: sections/introduction.tex
\vspace{-10pt}
\section{Introduction}

%\todo{Introduction - Manolis}

%zpp 3D reconstruction datasets enabled embodied AI, but limitations galore
Over the recent past, work on acquiring and semantically annotating datasets of real-world spaces has significantly accelerated research into embodied AI agents that can perceive, navigate and interact with realistic indoor scenes~\cite{anderson2018evaluation,savva2019habitat,shen2020igibson,szot2021habitat,ramakrishnan2021hm3d}.
However, the acquisition of such datasets at scale is a laborious process.
HM3D~\cite{ramakrishnan2021hm3d} which is one of the largest available datasets with 1000 high-quality and complete indoor space reconstructions, reportedly required 800+ hours of human effort to carry out mainly data curation and verification of 3D reconstructions.
Moreover, dense semantic annotation of such acquired spaces remains incredibly challenging.

%zpp we present largest semantically-annotated dataset; based on HM3D which is large and has complete, high-fidelity reconstructions
We present the Habitat-Matterport 3D Dataset Semantics (\hmtdsem).
This dataset provides a dense semantic annotation `layer' augmenting the spaces from the original HM3D dataset.
This semantic `layer' is implemented as a set of textures that encode object instance semantics and cluster objects into distinct rooms.
The semantics include architectural elements (walls, floors, ceilings), large objects (furniture, appliances etc.), as well as `stuff' categories (aggregations of smaller items such as books on bookcases).
This semantic instance information is specified in the semantic texture layer, providing pixel-accurate correspondences to the original acquired RGB surface texture and underlying geometry of the objects.
% \todo{Describe and motivate relative to HM3D~\cite{ramakrishnan2021hm3d} and MP3D~\cite{chang2017matterport3d}. We add semantics to match and exceed latter and show benefits of former for Embodied AI (ObjectNav) agent training.}

%zpp summary comparative statistics: more rooms, more objects, more categories; enable large-scale 3D vision, egocentric vision, and embodied AI experiments
The \hmtdsem dataset currently contains annotations for $\numobjects$ object instances distributed across $\numscenes$ spaces and $\numrooms$ rooms within those spaces. Figure~\ref{fig:teaser} shows some examples of the semantic annotations from the \hmtdsem dataset.
%Even though this is still only a part of the $1@000$ available scenes, 
The achieved scale is  larger than prior work (2.8x relative to Matterport3D~\cite{chang2017matterport3d} (MP3D) and 2.1x relative to ARKitScenes~\cite{dehghan2021arkitscenes} in terms of total number of object instances).
%zpp ObjectNav experiments; training on HM3DSem "subsumes" all other prior datasets
We demonstrate the usefulness of \hmtdsem on the ObjectGoal navigation task. 
Training on \hmtdsem results in higher cross-dataset generalization performance. Surprisingly, the policies trained on \hmtdsem perform better on average across scene datasets compared to training on the datasets themselves. We also show that increasing the size of training datasets improve the navigation performance. These results highlight the importance of improving the quality and scale of 3D datasets with dense semantic annotations for improving downstream embodied AI task performance.
% \todo{Summary of findings.}
%Prior work on ObjectNav in real-world scans has used a inconsistent set of object categories.
%The set of object categories for MP3D included objects that were poorly annotated (fireplace).
%\todo{Statistics on human performance on ObjectNav (MP3D vs \hmtdsem)}

%What is HM3D Semantics: object-level semantics for large-scale (building-level) indoor 3D environments.
%In addition to scale, we also provide high-quality texture level semantic annotation.  Texture level semantic annotations allows for more detailed segmentation of objects than point/vertex based annotations (cite scannet, scenenn?).  \todo{Cite some statistics that show our semantic annotation has less errors than prior work.}

%% file: sections/related.tex
\section{Related Work}

%\todo{Related work - Angel}
% \todo{Initial content copied from HM3D with some initial notes. Update to focus on the semantic aspect}

\xhdr{3D reconstruction datasets with semantics.}
There is a relatively small number of prior works that focus on semantically annotated 3D interior spaces acquired from the real world.
Collecting, reconstructing, and annotating such data at scale is a significant effort that requires complex pipelines and annotation tools.
Earlier work has therefore focused on scenes at the scale of single rooms.
For example, ScanNet~\cite{dai2017scannet} provided 707 typically room-scale reconstructions annotated with object semantic instances through labeling of 3D mesh segments constructed using an unsupervised segmentation algorithm.
Followup work by \citet{wald2019rio} adopted a similar approach and also targeted room-sized scenes.
Most recently, ARKitScenes~\cite{dehghan2021arkitscenes} contributed scans of 1661 room-scale scenes but only provides bounding box annotations for object instances.

Prominent prior works on building-scale datasets with semantic annotation are Matterport3D~\cite{chang2017matterport3d}, a subset of Gibson by \citet{armeni20193d}, and the Replica~\cite{straub2019replica} dataset.
The first uses the same methodology as ScanNet (labeling of 3D mesh segments), while the second provides human-verified object instance annotations created by back-projecting 2D semantic segmentation masks.
The third provides high-quality mesh vertex-level object instance labels but only contains 18 scenes.
Building on top of HM3D, which consists of over $1@000$ diverse environments from around the world, \hmtdsem provides detailed texture-level semantic annotations for building-scale reconstructions.

\xhdr{Synthetic 3D scene datasets.}
The use of synthetic 3D datasets for embodied AI simulation is quite common, especially when interactive environments are desired~\cite{kolve2017ai2,yan2018chalet,puig2018virtualhome,szot2021habitat}.
Due to the difficulty of modeling high-fidelity synthetic environments at scale, most existing datasets are limited in size and typically represent room-scale scenes.
Some of the prior work in this space has adopted a `teleportation' mechanism that allows an agent to immediately move from room to room through closed doors~\cite{yan2018chalet}.
A few datasets contributed by prior work focus on larger-scale scenes that coherently represent entire residences with multiple rooms~\cite{song2017semantic,fu20203dfront,szot2021habitat}.
These datasets have a number of limitations.
First, due to the difficulty in modeling a broad diversity of objects and scene layouts containing them, there is fairly limited variation in both object appearance and the spatial arrangements of the objects in the scenes.
Moreover, the objects exhibit modeling biases that create a simulation-to-reality gap, and the re-use of the same object models across scenes produces the unrealistic effect of ``perfect copies'' of particular objects.
These limitations have inspired work that attempts to tackle sim-to-real discrepancy by creating synthetic datasets that conform to scenes from the real world in terms of object appearance and spatial arrangement~\cite{avetisyan2019scan2cad,deitke2020robothor,szot2021habitat,li2021openrooms}.
However, this approach is hard to scale, and modeling biases due to the use of synthetic 3D data content creation software still remain.
In contrast, we focus on scaling high-quality semantic annotations of \emph{real} scenes acquired from a diverse set of spaces in the real world.

\setlength{\tabcolsep}{16pt}
\begin{savenotes}
% so we can have footnotes
\begin{table*}
\centering
{
\begin{tabular}{@{}lrrrrrrr@{}}
\toprule
Dataset & Scenes & Rooms & Object instances & Objects/room & Annotation type\\
\midrule
Replica~\cite{straub2019replica} & $18$ & $\approx25$ & $2@843$ & $\approx114$ & vertex\\
Gibson (tiny\footnote{Human-verified subset of Gibson~\cite{xiazamirhe2018gibsonenv} with semantic annotations.})~\cite{armeni20193d} & $35$ & $727$ & $2@397$ & $\approx3$ & vertex \\
ScanNet~\cite{dai2017scannet} &  $707$ & $\approx707$ & $36@213$ & $\approx24$ & segment \\
3RScan~\cite{wald2019rio} & $478$ & $\approx478$ & $43@006$ & $\approx29$ & segment \\
MP3D~\cite{chang2017matterport3d} & $90$ & $2@056$ & $50@851$ & $\approx25$ & segment \\
ARKitScenes~\cite{dehghan2021arkitscenes} & $1@661$ & $5@048$ & $67@791$ & $\approx13$ & bounding box \\
% \hmtdsem (ours) & $\numscenes$ & $\numrooms$ & $\numobjects$ & $\approx46$ & texture \\
\hmtdsem (ours) & $\numscenes$ & $\numrooms$ & $\numobjects$ & $\approx60$ & texture \\
\bottomrule
\end{tabular}
}
\vspace{-5pt}
\caption{
Comparison of \hmtdsem to other semantically annotated indoor scene datasets.
%Gibson 4+ refers to the subset of Gibson scenes that were rated as ``high quality'' and relatively free of reconstruction errors~\cite{savva2019habitat}.  
Statistics are on the publicly released portions of the corresponding datasets (does not include ScanNet or ARKitScenes hidden test sets).
% \todo{Add more appropriate statistics capturing scale, quality, and diversity of semantics?}
}
\label{tab:dataset-comparison}
\vspace{-10pt}
\end{table*}
\end{savenotes}

%% file: sections/dataset.tex
\section{Dataset Details}

% \todo{Dataset overview - Alex/Aaron} 

\input{fig/examples}

\begin{figure*}
    \centering
    \includegraphics[width=0.98\textwidth]{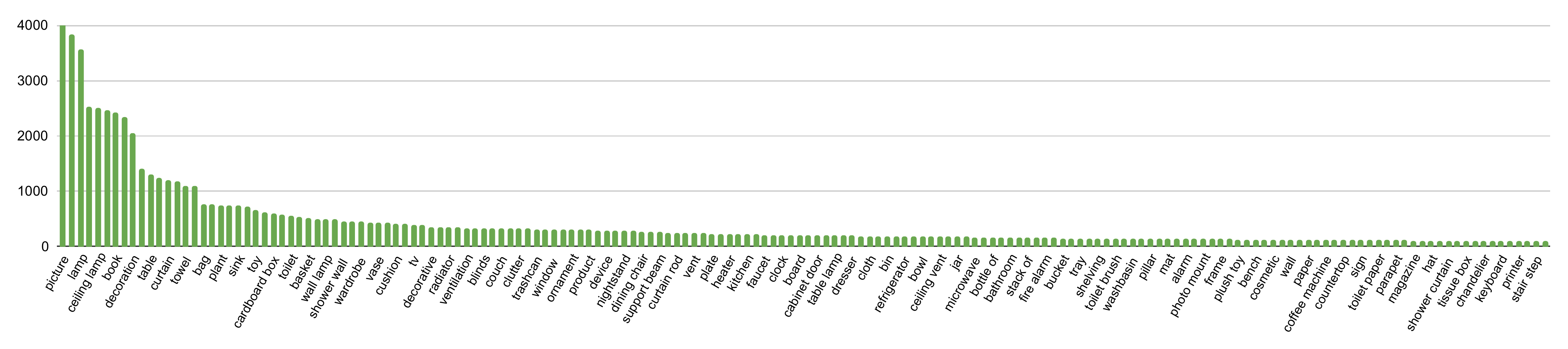}
    \vspace{-5pt}
    \caption{Histogram of most common semantic labels with 100+ instances across all scenes. Common architectural categories (e.g. floor, ceiling, wall) are not displayed.}
    \label{fig:top_100}
    \vspace{-15pt}
\end{figure*}

\begin{figure*}
\begin{minipage*}{\linewidth}
% \begin{figure}
    \centering
    \includegraphics[width=0.35\textwidth]{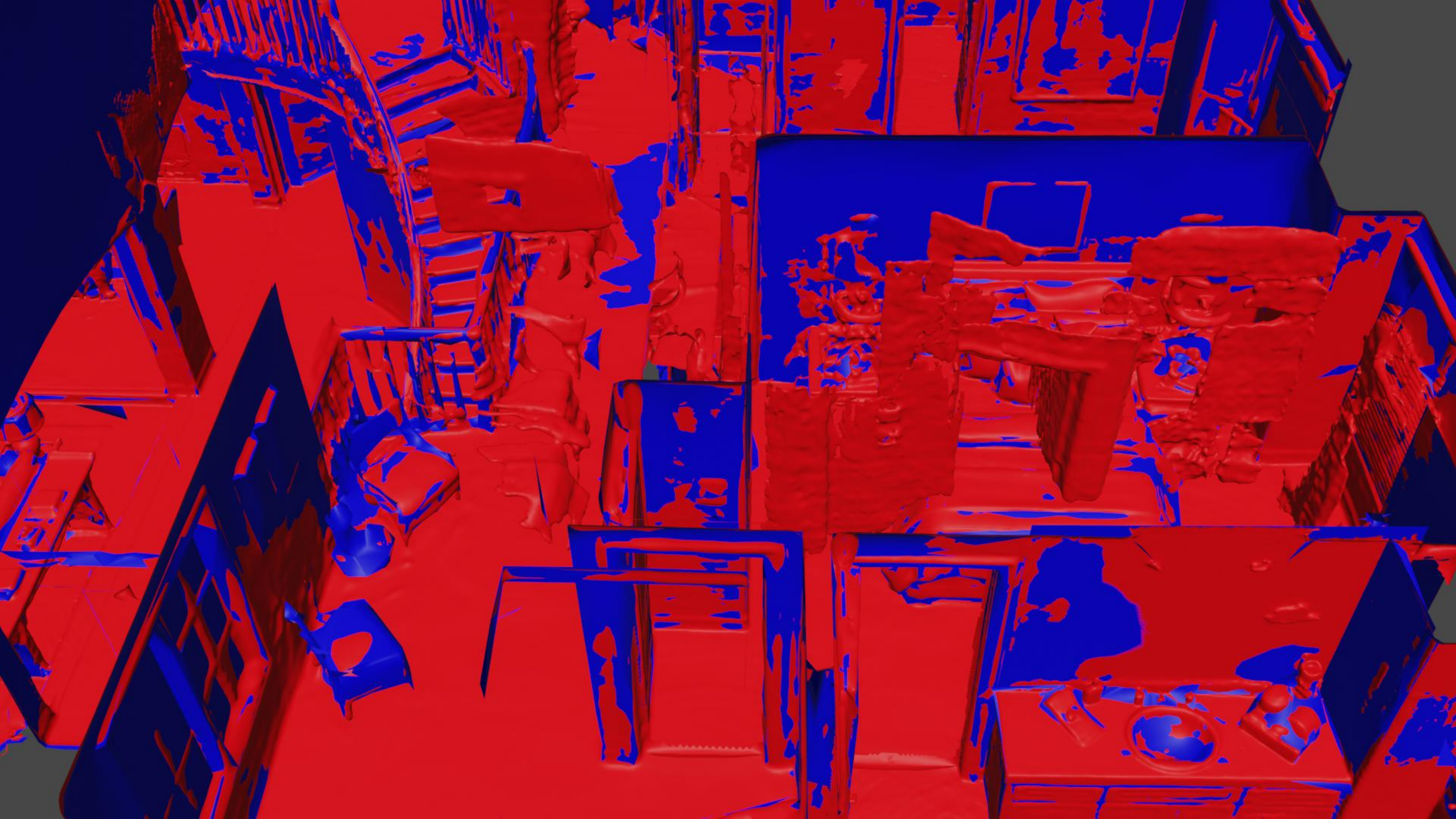}\hfill
    \includegraphics[width=0.35\textwidth]{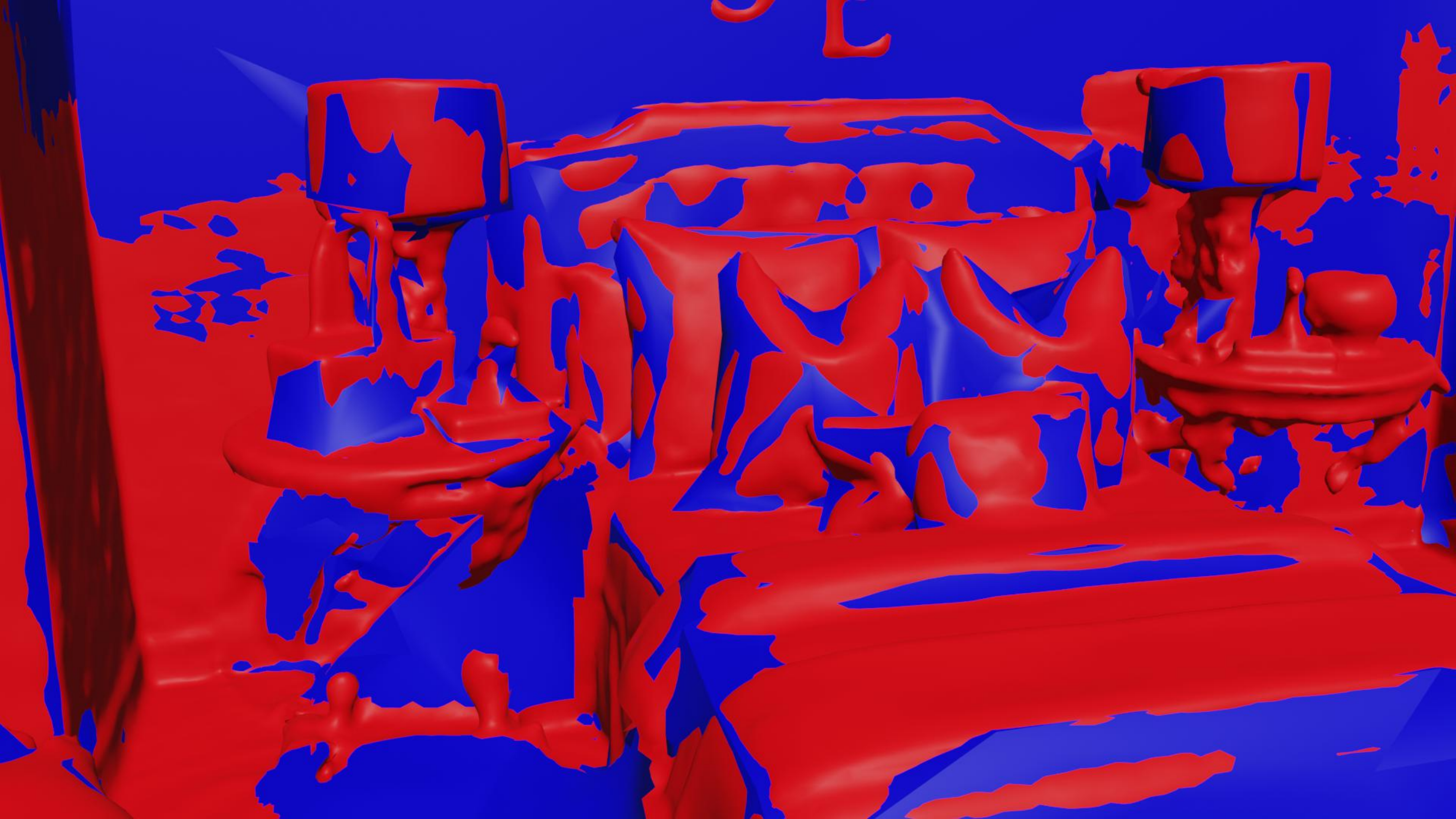}\hfill
    \includegraphics[width=0.22\textwidth]{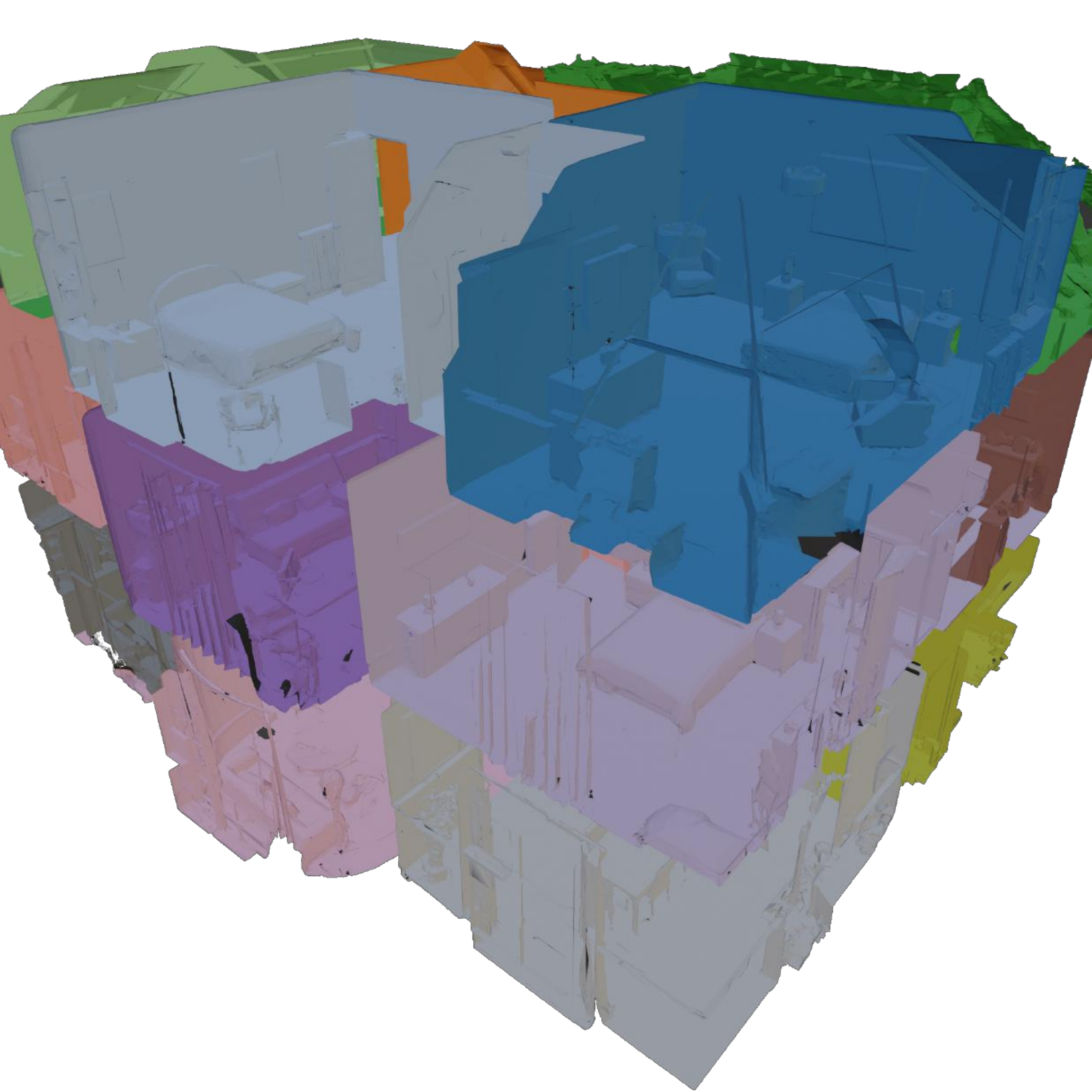}
    \vspace{-5pt}
    \captionof{figure}{Examples of major (\textbf{left}) and minor (\textbf{center}) misalignment between the scanned RGB mesh (blue) and auto-generated semantic mesh (red) from the MP3D dataset \cite{chang2017matterport3d}. \textbf{Right}: Visualization of region annotations in \hmtdsem. Each color is an aggregation of all instances mapped to a particular region.}
    \label{fig:mp3d_errors}
    \vspace{-10pt}
    % 
    % \captionof{figure}{}
    % \label{fig:region_annotations}
% \end{figure}
\end{minipage*}
\end{figure*}

% \alex{Mostly copied from HM3D Semantics webpage.}
The Habitat-Matterport 3D Semantics Dataset is the largest-ever human-annotated dataset of semantically-annotated 3D indoor spaces.
%\alex{Is this true? Need to specify "human-annotated" right?} 
It contains dense semantic annotations for $\numscenes$ high-resolution, 3D, scanned scenes from the Habitat-Matterport 3D Dataset (HM3D). The HM3D scenes are annotated with $\numobjects$ raw object names additionally mapped to the 40 Matterport 3D categories \cite{chang2017matterport3d}. On average, each scene consists of $661$ objects from $106$ categories. This dataset is the result of over 14,200
hours of human effort for annotation and verification by 20+ annotators. The following subsections provide further details on asset formats, the annotation pipeline, and scene content statistics.

\subsection{Data Format and Contents}
The semantic annotations are available as a set of texture images applied to the original scene geometry from HM3D and packed into binary glTF (.glb) format. Unique hex colors differentiate each object instance and map it to a raw text string classifying the instance. These mappings are included in a metadata text file accompanying the .glb asset, which additionally labels each instance with a region ID to define object grouping by room. 

Often, semantic annotations are defined per-vertex and directly embedded in the mesh geometry (e.g., ScanNet~\cite{dai2017scannet}, Gibson~\cite{shen2020igibson}, and MP3D~\cite{chang2017matterport3d}).
However, it is not uncommon for mesh geometry discretization to insufficiently capture boundaries between objects, especially on flat surfaces such as walls, floors, and table-tops. This results in jagged inaccurate semantic boundaries, missing annotations, or requires generating an entirely new mesh with higher resolution than the original, which has implications on both rendering performance and visual alignment. For example, \Cref{fig:mp3d_errors} highlights the common misalignment errors between annotated and original assets from the MP3D dataset resulting from automated mesh geometry generation. In contrast, \hmtdsem archival format encodes annotations directly in a set of textures compatible with the original geometry. As it is not uncommon for 3D assets, especially those derived from scanning pipelines to represent object boundaries in texture rather than geometry, this choice seemed natural. \Cref{fig:examples} shows several example scenes and contrasts them against semantic annotations from Matterport3D~\cite{chang2017matterport3d}, which is the most related prior dataset. The density and quality of semantic instance annotations in \hmtdsem exceeds that of prior work as shown in \Cref{tab:dataset-comparison}. For additional compatibility with existing simulators, the semantic texture annotations are also baked into per-vertex colors included with the assets.

% Semantic annotations were performed manually by artists using 3D modeling software. This process was completed via a texture painting pipeline wherein unique instance colors were painted directly onto UV mapped scene geometry to fill initially empty texture images. %\todo{Bring up more details about contributions across artists from the breakdown.}

Artists were instructed to annotate architectural features such as: walls, floors, ceilings, windows, stairs, and doors as well as notable embellishments such as door and window frames, banisters, area rugs, and moulding. Instance annotations for architectural features are broken into regions at transition points such as room boundaries, doorways, and hallways to more readily classify components into regions (e.g. to semantically separate floors and ceilings as a room transitions to a hallway) as shown in \Cref{fig:mp3d_errors} (right). Additionally, decorative features such as pictures, posters, switches, vents, lighting fixtures, and wall art are segmented and labeled.

Furniture, appliances, and clutter objects were annotated and segmented from their surroundings whenever possible. For example, pillows and blankets are segmented individually from beds, couches, and chairs while remote controls, electronics, lamps, and art pieces are segmented from desks, tables, and consoles. In many cases, as scan resolution permits, individual clothing items, linens, and books are segmented from one another in closets and bookshelves.

\subsection{Verification Process}

Annotation on the scale of HM3D Semantics is not a one-way street. Roughly 640 annotator hours were allocated to iteration and error correction (about 4.5\% of all annotator hours). Additional verification was done by the authors, including both qualitative manual assessment and automated programmatic checks. Even so, some errors may yet remain. Fortunately, the archival format of texture + text allows for efficient iterative improvement of the annotations. 

Automated verification is essential for large scale annotation efforts. Our automated verification pipeline included, among others, the following checks:
\begin{itemize}
    % \item Texture images contain only annotated colors. \todo{We check, but haven't enforced this.}
    \item Text file annotations contain only colors from textures.
    \item Each annotation color used only once per scene.
    \item Text file contents conform to expected format: index, color, category name, region id.
\end{itemize}

Qualitative verification proves challenging to automate, and as such, manual validation by humans remains an important part of the annotation QA pipeline. Following delivery of the annotated assets, a manual review and iteration phase was conducted, including the following:
\begin{itemize}
    \item Validation pass over raw text names included identification and correction of typos, consolidation of synonyms, and mapping of raw text names to the 40 canonical object classes from the MP3D dataset \cite{chang2017matterport3d}. 
    \item Visual inspection through virtual walk-through in Habitat~\cite{szot2021habitat}. Verifiers checked for missing annotations, messy boundaries, annotation artifacts, over-aggregation (i.e., multiple unique instances sharing an annotation color), semantic mislabeling (e.g. ``dishwasher'' annotated as ``washing machine''), and other common flaws.
\end{itemize}

\subsection{Dataset Statistics}

The \numscenes scenes chosen as candidates for \hmtdsem annotation were selected at random from the 950 furnished HM3D scan assets. These are distributed into subsets of [145, 36, 35] scenes between [train, val, test] splits.
The 36th val model is an example scene freely available without registration for quick inspection and automated testing of downstream dataset use cases.

An analysis of the annotation text files reveals much about the contents of the scanned environments. There are $\numrawcategories$ category tags labeling the $\numobjects$ object instances across the entire dataset, split amongst $\numrooms$ regions. Each scene contains, on average, $106$ unique categories, $660$ object instances, and $14$ annotated regions. The histograms in \Cref{fig:contents_histograms} show the overall distribution of regions, object instances, and unique categories across all scenes in the dataset. It is worth noting that because annotators were given freedom when defining category tags, many synonymous tags are present in the final dataset.

Of the $\numobjects$ object instances present in the dataset, $34@368$ are either labeled as “unknown” or belong to architectural categories, such as “wall”, “door”, “ceiling”, etc, leaving approximately $108@278$ annotated object instances. Those categories with 100 or more instances throughout the dataset are shown in \Cref{fig:top_100}.

Each annotated region contains on average, $46$ object instances and $20$ unique categories. Further statistical observations were made using a region labeling heuristics where proposed region/room labels were derived from object category-inferred proposals. For example the presence of a “bed” instance implies that the containing region is a “bedroom” and a “toilet” implies a “bathroom”. See the supplemental material for more details and source data sheets.

\begin{figure}
    \centering
    \includegraphics[width=0.49\textwidth]{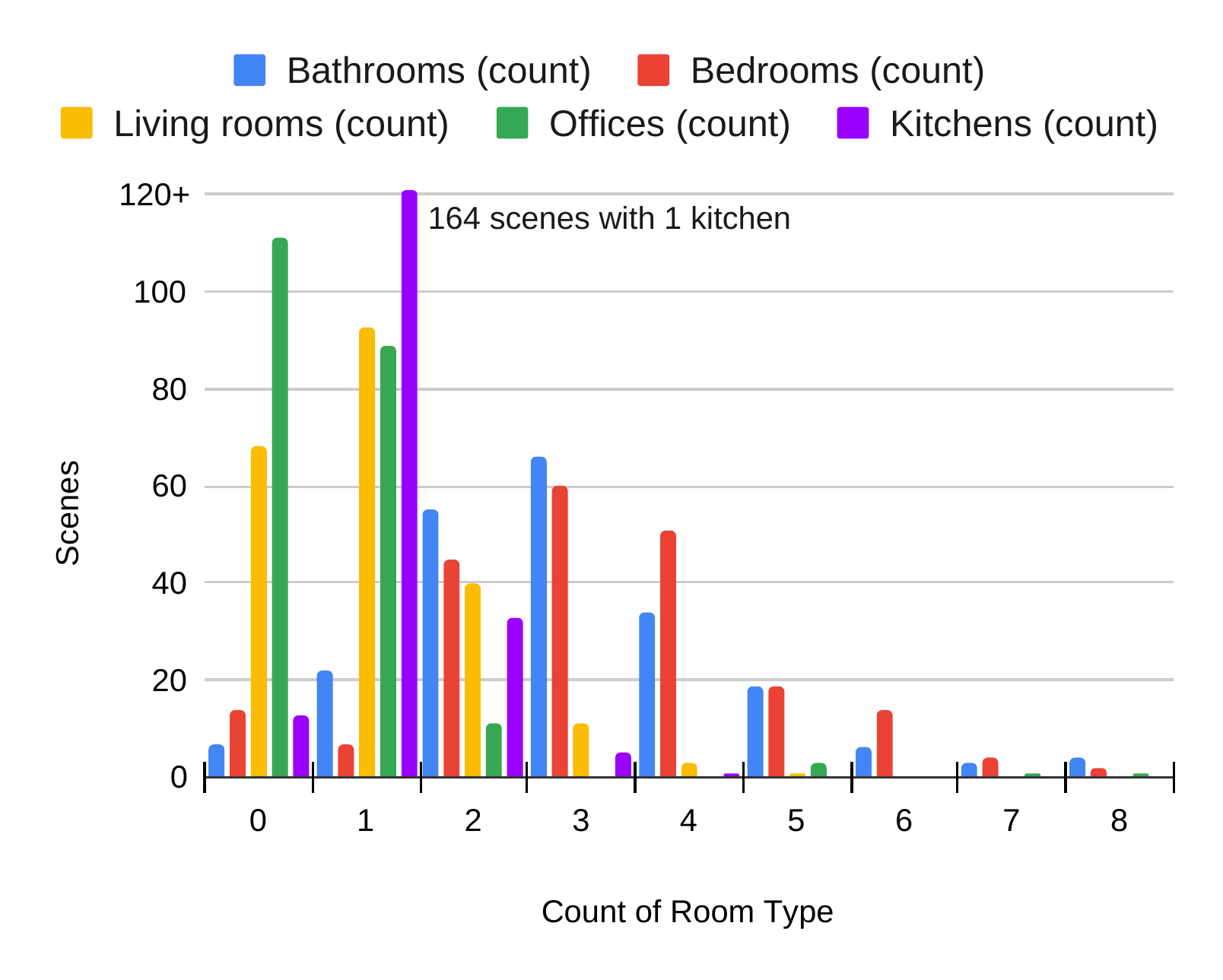}
    \vspace{-25pt}
    \caption{Histogram of room types in each scene based on region instance mapping and category implied room type heuristic. Vertical axis clamped to 120.}
    \label{fig:room_types_histogram}
    \vspace{-15pt}
\end{figure}

Given regions labeled using these heuristics, we can investigate the prevalence of individual room types within the scenes and cluster them by their expected contents. \Cref{fig:room_types_histogram} shows a histogram of common room types counted per-scene. From these statistics we can see that:
\begin{itemize}
 \item More than 30\% of scenes are larger residences with 4+ bedrooms and bathrooms.
 \item Many scenes have more than 1 kitchen, possibly indicating multi-family homes or multiple individual living units packed into a single scene.
 \item A small set of very large scenes have 5+ regions labeled as offices and living rooms.
\end{itemize}

Further analysis of the raw data reveals: 
\begin{itemize}
    \item 14 scenes lacked any heuristically labeled "bedroom" regions. These scenes were all visually verified to be commercial spaces such as offices, restaurants, or stores. 
    \item 7 scenes lacked any “bathroom” regions. These were also visually verified to be non-residential spaces (a subset of the 14 which lacked "bedroom" labels). 
    \item 25 scenes contained “garage” regions. These were visually verified to all contain garages.
    \item 13 scenes lacked any “kitchen” regions. 5 of these were commercial spaces that also lacked “bedroom” labels, while 5 of the remaining 6 were hotel rooms or suites. The final scene contained a kitchen through visual inspection, however the modern design lacked most obvious appliances and was therefore not heuristically labeled as such.
\end{itemize}

We hope these statistical insights enable researchers to pick specific subsets of scenes for their experiments based on relevant criteria. Additionally, further statistical analysis of these data may reveal deeper relationships between objects and their common regions or neighborhoods. The results of this analysis may be useful in downstream tasks such as scene understanding and procedural generation.

% \begin{figure}[t!]
    % \centering
    % \includegraphics[width=0.3\textwidth]{fig/region_annotation_example.png}
    % \caption{Visualization of region annotations. Each color is an aggregation of all instances mapped to a particular region.}
    % \label{fig:region_annotations}
    % \vspace{-10pt}
% \end{figure}

%% file: fig/examples.tex
\begin{figure*}
% \centering
% \setkeys{Gin}{width=\linewidth}
% \begin{tabularx}{\textwidth}{Y Y Y Y Y Y}
% % \imgclip{0.3}{0.9}{fig/qual/hm3d/1S7LAXRdDqK.basis.png}
% \multicolumn{3}{c}{\hmtdsem} & \multicolumn{3}{c}{MP3D}\\
% \cmidrule(lr){1-3} \cmidrule(lr){4-6}

% \includegraphics{fig/qual/hm3d/1S7LAXRdDqK.basis.png} &
% \includegraphics{fig/qual/hm3d/8B43pG641ff.basis.png} &
% \includegraphics{fig/qual/hm3d/iePHCSf119p.basis.png} &
% \includegraphics{fig/qual/mp3d/17DRP5sb8fy.faceids-0.png} &
% \includegraphics{fig/qual/mp3d/2t7WUuJeko7.faceids-0.png} &
% \includegraphics{fig/qual/mp3d/GdvgFV5R1Z5.faceids-0.png}
% \\
% \includegraphics{fig/qual/hm3d/1S7LAXRdDqK.semantic.png} &
% \includegraphics{fig/qual/hm3d/8B43pG641ff.semantic.png} &
% \includegraphics{fig/qual/hm3d/iePHCSf119p.semantic.png} &
% \includegraphics{fig/qual/mp3d/17DRP5sb8fy.faceids-0.objectId.png} &
% \includegraphics{fig/qual/mp3d/2t7WUuJeko7.faceids-0.objectId.png} &
% \includegraphics{fig/qual/mp3d/GdvgFV5R1Z5.faceids-0.objectId.png}
% \\
% \end{tabularx}
\centering
\includegraphics[width=0.88\linewidth]{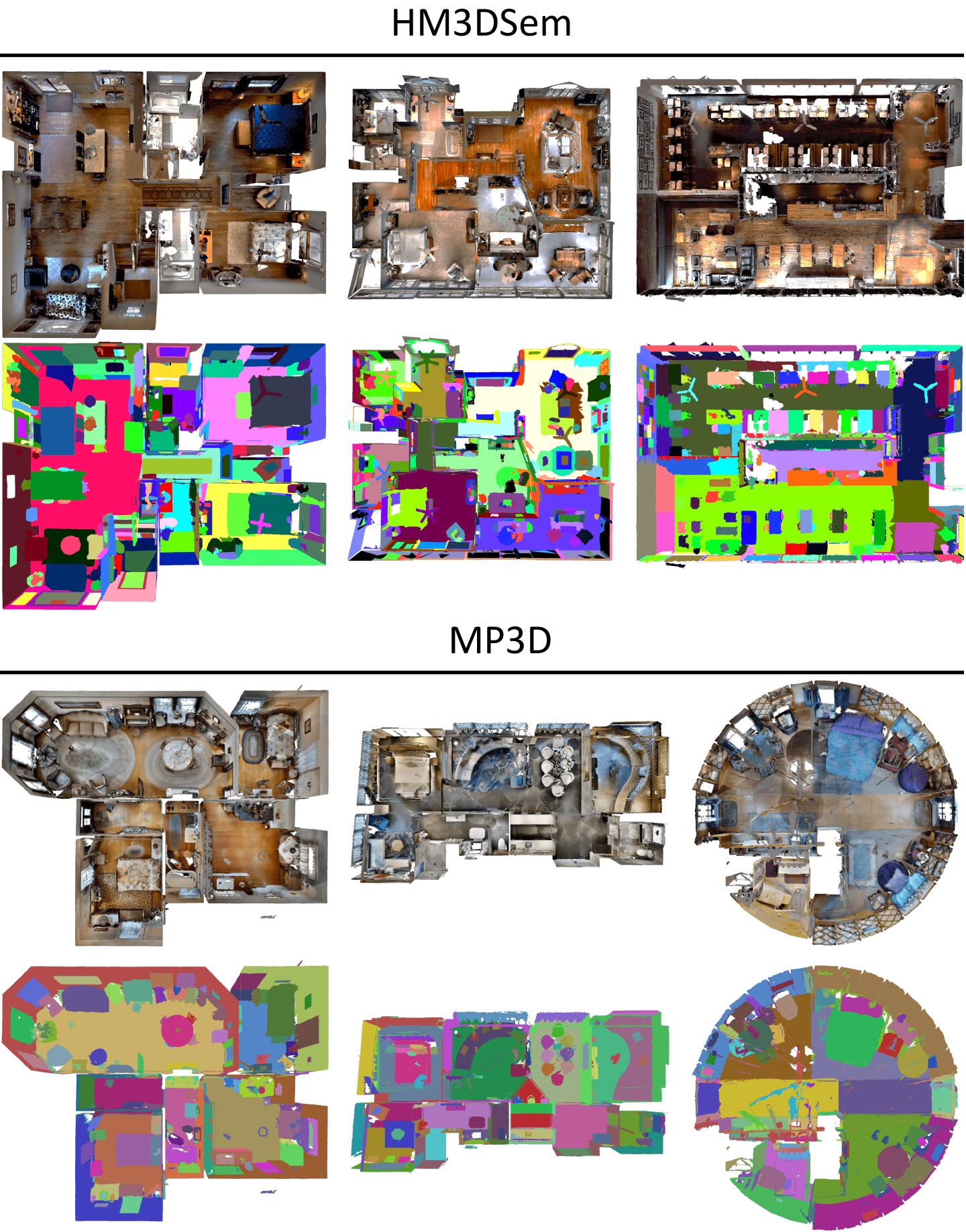}
\caption{Qualitative examples comparing semantic annotations of scenes from \hmtdsem (top) and Matterport3D~\cite{chang2017matterport3d} (bottom).
The first row in each pair of rows shows a top-down view of the scene.
The second row shows semantic object instances in distinct colors.
The \hmtdsem annotations provide a greater number of distinct object instances, as indicated also by the summary statistics in \Cref{tab:dataset-comparison}).
Many paintings and other wall objects are annotated in \hmtdsem (see leftmost wall in top right scene).
Smaller object types such as decorative pieces on bookcases (see top left scene, leftmost corner) are also annotated.
In contrast, semantic annotations from Matterport3D tend to cluster smaller objects into larger furniture pieces (e.g., items on piano at top left, and items on nightstand and bed in rightmost scene).
}
\label{fig:examples}
\end{figure*}

%% file: sections/experiments.tex
\section{Experiments}

In this section, we present experimental results for training Object-Goal Navigation (ObjectNav) policies using the \hmtdsem dataset in the Habitat simulator~\cite{savva2019habitat}. To compare the quality of \hmtdsem with prior datasets, we train three different policies (reinforcement, imitation and modular learning) for ObjectNav using three different datasets, \hmtdsem, Gibson~\cite{shen2020igibson}, and MP3D~\cite{chang2017matterport3d}. We then evaluate each policy on all datasets. For example, a policy trained on \hmtdsem will be evaluated on Gibson and MP3D, even though it was not trained on them. We show that the policies trained on \hmtdsem perform better or are comparable to those trained on Gibson and MP3D when evaluated on all three datasets. This indicates that training ObjectNav policies on \hmtdsem improves cross-dataset domain generalization. We also show that increasing the number of scenes used for training leads to better generalization to previously unseen scenes.

\textbf{ObjectNav task definition.} For our experiments we use an agent matching LoCobot's specification with a base radius of 0.18m and height of 0.88m. The agent is equipped with a 640x480 RGB-D camera (mounted at a height of 0.88m) along with a Compass and a GPS sensor. The agent's action space comprises of the [\code{MOVE\_FORWARD}, \code{TURN\_LEFT}, \code{TURN\_RIGHT}, \code{LOOK\_UP}, \code{LOOK\_DOWN}, \code{STOP}] actions with a forward step of $0.25\si{m}$ and turn angles of $30^\circ$. We define 6 goal categories (similar to~\cite{chaplot2020object}): chair, bed, plant, toilet, tv/monitor, and sofa. The agent is successful if it executes \code{STOP} at a location that lies within $1.0\si{m}$ of any object instance from the goal category. While the agent need not directly see the object while stopping, we require that the object can be directly viewed from the stop location without obstruction (i.e., \emph{oracle-visibility}~\cite{batra2020objectnav}). We evaluate the agent's performance using the standard Success and SPL metrics~\cite{anderson2018evaluation}. Success measures how often the agent finds and stops at the goal object, while SPL measures how efficiently the agent succeeds (i.e., the efficiency of the agent's path relative to the shortest path from the start to goal positions).

\textbf{ObjectNav episode dataset.} We generate episode datasets from the $\numtrainscenes$ train, $\numvalscenes$ val, and $\numtestscenes$ test scenes for benchmarking agents on the ObjectNav task. Our episode generation process is similar to prior ObjectNav work~\cite{batra2020objectnav}. Each episode consists of a scene, a start position where the agent is placed at time $t=0$, and a goal object category. To generate an episode for a given scene, we uniformly sample a goal from the 6 goal categories. We then randomly sample a start location from the scene that satisfies the following constraints: (1) the start location must be navigable, (2) the goal object must be reachable from the start location, and (3) the distance from the start location to the nearest object from the goal category must lie between $1\si{m}$ and $30\si{m}$. Following this procedure, we generate episode datasets containing $\sim7.2\si{M}$ train / $1072$ val / $1000$ test episodes.

\input{sections/experiments-main-table}

\begin{figure}
    \includegraphics[width=0.49\textwidth]{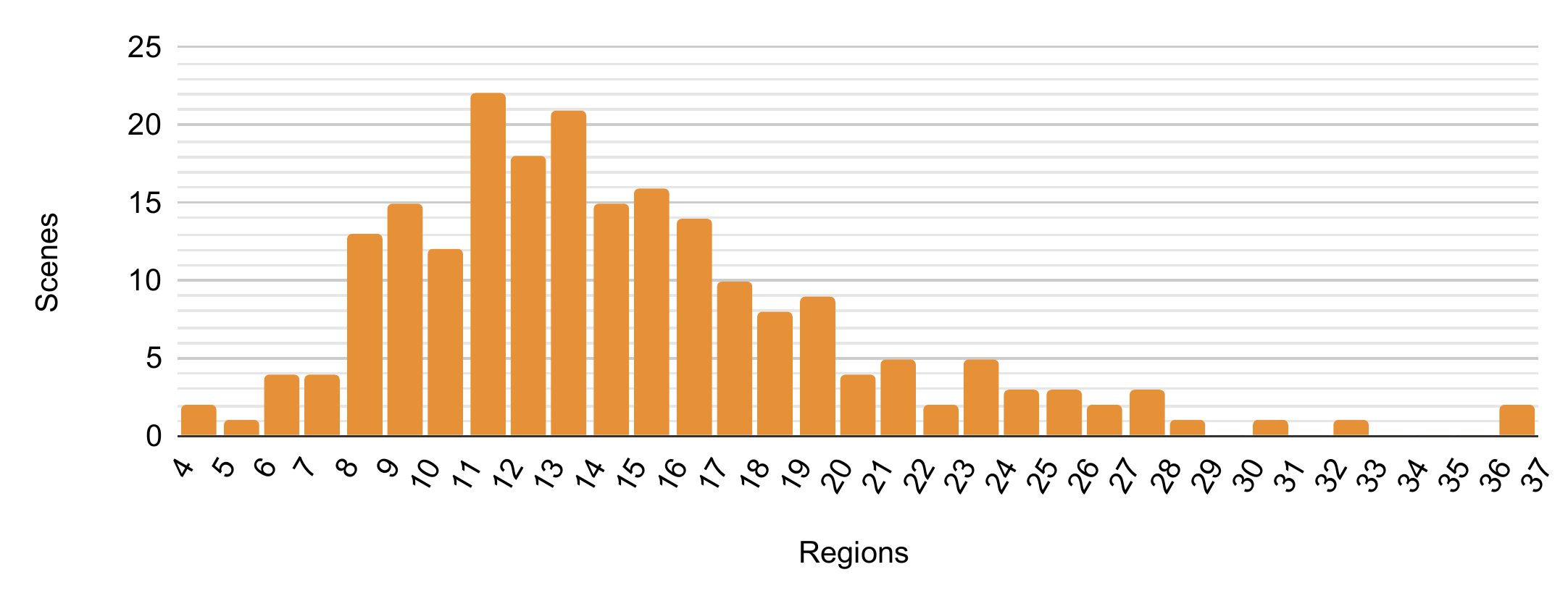}
    \includegraphics[width=0.49\textwidth]{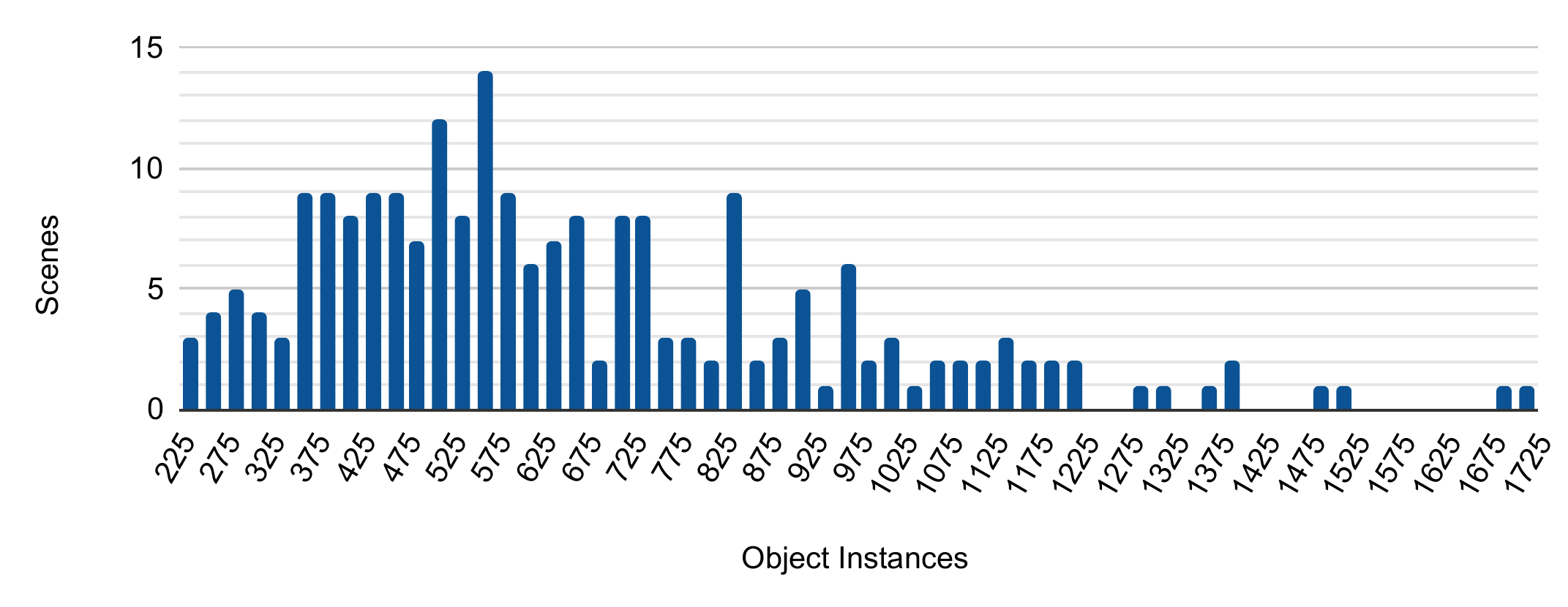}
    \includegraphics[width=0.49\textwidth]{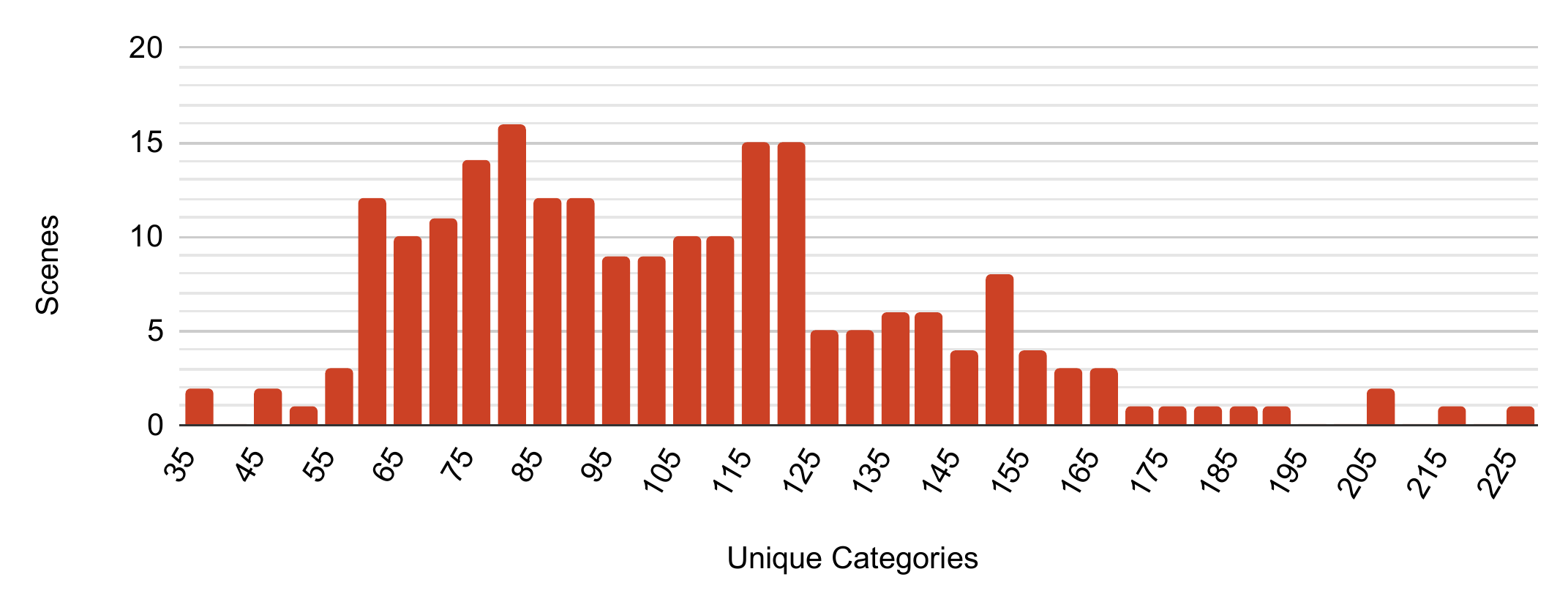}
    \vspace{-20pt}
    \caption{Histograms displaying region, object instance, and unique annotation category counts over all scenes in the dataset.}
    \label{fig:contents_histograms}
    \vspace{-10pt}
\end{figure}

\setlength{\tabcolsep}{6pt}
\begin{table}
\begin{minipage}{\linewidth}
    \centering
    
    \resizebox{0.85\textwidth}{!}{%
    \begin{tabular}{@{}cl|c|c|cc@{}}
    \toprule
    &                  &              &        & \multicolumn{2}{c }{\hmtdsem (val)} \\
    & Name    & Area (m$^2$) & Scenes & Success $\uparrow$ &  SPL $\uparrow$    \\ \midrule
    \parbox[t]{2mm}{\multirow{4}{*}{\rotatebox[origin=c]{90}{\small{\hmtdsem}}}} 
    & Tiny    & 3811.76      &   25   & 17.79 & 6.70 \\
    & Small   & 9194.19      &   80   & 26.87 & 11.18 \\
    & Medium  & 12427.08     &   99   & 35.16 & 14.90 \\
    & Large   & 16523.37     &   145  & 39.36 & 18.02 \\ \bottomrule
    \end{tabular}
    }
    \vspace{-5pt}
\caption{\small \textbf{Dataset size vs Performance. (RL)} ObjectNav performance of an RL policy with increasing training dataset sizes.}
    \label{tab:objectnav_dataset_scale}
    \vspace{-10pt}
\end{minipage}
\end{table}

\subsection{Reinforcement Learning} 
\label{sec:rl_expts}
%\todo{Karmesh}
\textbf{Training Details.} We use the architecture proposed in DDPPO \cite{wijmans2019dd} to train the ObjectNav agents. A ResNet50 network encodes the RGB-D images into visual representations, which are concatenated with the embeddings of the goal object category, the previous action, and the GPS+Compass sensor readings. This joint embedding is passed to a 2-layer 512-D LSTM network. The output of the LSTM module is passed to fully-connected layers, which predict the action probabilities and state values. The agent is trained for 400M steps, after which it overfits on the training dataset. 

\textbf{Scene Dataset Comparison.} The top 3 rows in \Cref{tab:objectnav_datasets} show the performance of DDPPO agents trained on the Gibson, MP3D, and \hmtdsem episodes. We evaluate these agents on the validation scenes from all three datasets. We observe that training on \hmtdsem leads to the best performance averaged over all datasets. Surprisingly, the agent trained on \hmtdsem outperforms the agent trained on Gibson when evaluated on Gibson. This indicates that improved visual reconstruction and annotation accuracy in \hmtdsem leads to policies that generalize better even across datasets.

\textbf{Dataset Size vs Performance.} We also train agents on different subsets of \hmtdsem scenes to study the effects of dataset scaling. In \Cref{tab:objectnav_dataset_scale}, we observe that the performance of our agent improves with more training scenes, with \hmtdsem-Large performing more than twice as good as \hmtdsem-Tiny (39.36\% vs 17.70\%).

\subsection{Imitation Learning} 

%\todo{Ram}

% Compress this version
\textbf{Training Details}. We collect $77k$ human demonstrations for $80$ \hmtdsem training scenes using Habitat-Web~\cite{ramrakhya2022}.
Following \cite{ramrakhya2022} we use a simple CNN+RNN architecture.
For RGB, we use a ResNet18~\cite{he_cvpr16} that is randomly initialized.
For depth, we use a ResNet50 which was pre-trained on PointGoal navigation task using DD-PPO~\cite{wijmans2019dd} on gibson dataset.
The GPS+Compass inputs are passed through fully-connected layers to embed them to $32$-D vectors.
In addition to RGB-D and GPS+Compass, we use two additional semantic features following~\cite{ye_iccv21} -- semantic segmentation of the input RGB observation, predicted using a RedNet~\cite{jiang2018rednet} model,
and a `Semantic Goal Exists' feature which is the total area of the visual input occupied by the goal object category.
To predict the semantic features, we use the RedNet model from~\cite{ye_iccv21}, which was pre-trained on SUN RGB-D~\cite{song2015sun} and finetuned on $100k$ randomly sampled views from MP3D scenes.
Finally, we also feed in the object goal category embedded into a $32$-D vector.
These input features are concatenated to form an observation embedding,
and fed into a $2$-layer, $2048$-D GRU at every timestep.
We train this policy for ${\sim}400$M steps, which amounts to ${\sim}20$ epochs on ${\sim}77k$ demonstration episodes.

% \setlength{\tabcolsep}{10pt}
% \begin{table*}
% \begin{minipage}{\linewidth}
%     \centering
%     {%
%     \begin{tabular}{@{}l|cc|cc|cc@{}}
%     \toprule
%     & \multicolumn{2}{c|}{Gibson (val)} & \multicolumn{2}{c|}{MP3D (val)} & \multicolumn{2}{c }{HM3D (val)} \\
%      Dataset                 & Success $\uparrow$ &  SPL $\uparrow$  &Success $\uparrow$ & SPL $\uparrow$   & Success $\uparrow$ & SPL $\uparrow$    \\ \hline
%     MP3D                    & 18.95 & 6.95 & $\mathbf{28.57}$ & $\mathbf{9.91}$ & 25.37 & 7.73 \\
%     Gibson-Tiny             & 18.15 & 7.84 & 16.57 & 5.84 & 20.06 & 6.03 \\
%     HM3D                    &$\mathbf{28.02}$ &$\mathbf{11.64}$ & $25.06$ & $8.72$ &$\mathbf{33.49}$ &$\mathbf{11.64}$ \\ \bottomrule
%     \end{tabular}
%     }
%     \vspace{-5pt}
%     \caption{\small \textbf{Imitation Learning results.} ObjectNav performance of a Behavior Cloning policy when trained and tested on different datasets. We report the mean and standard deviation by training on 1 random seed, and evaluating on 3 random seeds. \vspace{-0.2cm}}
%     \label{tab:objectnav_table} 
% \end{minipage}
% \end{table*}

\setlength{\tabcolsep}{6pt}
\begin{table}
\begin{minipage}{\linewidth}
    \centering
    
    \resizebox{\textwidth}{!}{%
    \begin{tabular}{@{}cl|c|c|cc@{}}
    \toprule
    &                  &              &        & \multicolumn{2}{c }{\hmtdsem (val)} \\
    & Name    & Area (m$^2$) & Scenes & Success (\%) $\uparrow$ &  SPL (\%) $\uparrow$    \\ \midrule
    \parbox[t]{2mm}{\multirow{4}{*}{\rotatebox[origin=c]{90}{\small{\hmtdsem}}}} 
    & Tiny-HD    & 3811.76      &   25   & 34.98 & 11.84 \\
    & Small-HD   & 5753.60      &   36   & 38.90 & 13.19 \\
    & Medium-HD  & 7099.31      &   54   & 48.32 & 16.07 \\
    & Large-HD   & 9194.19      &   80   & 54.20 & 18.71 \\ \bottomrule
    \end{tabular}
    }
    \vspace{-5pt}
\caption{\small \textbf{Dataset size vs Performance. (IL)} ObjectNav performance of an IL policy with increasing human demonstrations dataset sizes for training.}
    \label{tab:objectnav_il_dataset_scale}
    \vspace{-15pt}
\end{minipage}
\end{table}

\textbf{Scene Dataset Comparison}. \Cref{tab:objectnav_datasets} (rows 4-6) shows the performance of ObjectNav policies trained using imitation learning (specifically, behavior cloning) on $10k$ human demonstrations from the \hmtdsem, Gibson and MP3D datasets.
%We evaluate each policy on other unseen datasets in a zero-shot fashion, i.e. a policy trained on \hmtdsem train scenes will be evaluated on MP3D and Gibson validation scenes without any training on the corresponding train scenes.
 We evaluate these agents on the validation scenes from all three datasets.
We find that the imitation learning policy trained on \hmtdsem achieves the best performance averaged across all validation datasets. In fact, the \hmtdsem policy even outperforms the Gibson policy when evaluated on Gibson validation scenes. This echoes our findings from~\cref{sec:rl_expts} and further emphasizes the high annotation quality in \hmtdsem.

%%%%%%% @Ram --- Uncomment if all these are required %%%%%%%%%%%
%We find that a policy trained on \hmtdsem outperforms policies trained on Gibson and MP3D, when evaluated on validation episodes from Gibson and \hmtdsem (results in col $1$ and col $3$).
%Similarly, a policy trained on \hmtdsem outperforms the policy trained on Gibson when evaluated on MP3D validation (results in col $2$) and performs $3.51\%$ worse compared to a policy trained on MP3D scenes. 
%Overall, we find that a policy trained on \hmtdsem scenes generalizes well to unseen scene datasets compared to policies trained on Gibson and MP3D scenes.

\textbf{Dataset Size vs Performance}. In~\Cref{tab:objectnav_il_dataset_scale}, we study the dataset scaling behavior by training on different subsets of \hmtdsem scenes, ranging from 25 to 80 scenes. 
%We created scene splits for $80$ scenes used to collect human demonstrations dataset of increasing size from $25$ to $80$ scenes and trained policies with same hyperparameters on each split.
%
% Increasing the number of training scenes consistently increases the performance on the validation split, suggesting that it is valuable to scale the number of scenes for collecting additional demonstrations, and will likely improve the performance further.
We observe consistent improvements in the validation performance as we increase the number of training scenes. This suggests that it is valuable to collect large-scale human demonstrations for ObjectNav and that the performance is likely to improve further as we increase the number of training scenes.

% \begin{figure}[t]
%     \centering
%     \includegraphics[width=0.95\linewidth]{fig/IL/hm3d_dataset_size_vs_perf_v2.png}
%     \vspace{-6pt}
%     \caption{Success on ObjectNav \hmtdsem-\textsc{val} split \vs number of human demonstrations for training.}
%     \label{fig:dataset_size_vs_perf}
%     \vspace{-10pt}
% \end{figure}

% \begin{table} %[h!]
%     \centering
%     \resizebox{1\linewidth}{!}{
%         \begin{tabular}{@{}cllrrcrrrrr@{}}
%             \toprule
%             & & & \multicolumn{2}{c}{\textsc{HM3D}} & & \multicolumn{2}{c}{\textsc{MP3D}} & & \multicolumn{2}{c}{\textsc{Gibson}} \\
%             \cmidrule{4-5} \cmidrule{7-8} \cmidrule{10-11}
%             & & Method & Success $\%$ $(\mathbf{\uparrow})$ & SPL $\%$ $(\mathbf{\uparrow})$
%                 & & Success $\%$ $(\mathbf{\uparrow})$ & SPL $\%$ $(\mathbf{\uparrow})$ 
%                 & & Success $\%$ $(\mathbf{\uparrow})$ & SPL $\%$ $(\mathbf{\uparrow})$ \\
%             \midrule
%             \\[-10pt]
%             & \rownumber & IL w/ 10k Human Demos
%                 & $34.7$ & $14.1$
%                 & & $25.7$ & $8.0$
%                 & & $18.1$ & $7.8$\\
%             & \rownumber & Humans
%                 & $84.8$ & $46.5$
%                 & & - & -
%                 & & $86.4$ & $47.6$\\[5pt]
%             \bottomrule
%         \end{tabular}
%     }
%     \vspace{5pt}
%     \caption{ObjectNav IL results on HM3D, MP3D and Gibson ObjectNav VAL split.}
%     \label{tab:onav_dataset_results}
% \end{table}

\subsection{Modular Learning}
In addition to end-to-end reinforcement and imitation learning, modular learning has emerged as a popular alternative for training policies to tackle various Embodied AI tasks~\cite{chaplot2020learning, chaplot2020neural, chaplot2020semantic, chaplot2020object, Krantz_2021_ICCV, chaplot2021seal, georgakis2021learning, hahn2021nrns, min2021film, sarch2022tidee, ramakrishnan2022poni}. Besides showing that training on \hmtdsem leads to better end-to-end navigation policies, we also show that it leads to better modular components. Specifically, we train the Goal-Oriented Semantic Exploration (SemExp) policy of \cite{chaplot2020object} on \hmtdsem, Gibson, and MP3D and evaluate its generalization to other datasets.

\textbf{Training Details.} The approach proposed in \cite{chaplot2020object} builds a top-down semantic map by projecting first-person semantic segmentation predictions with depth, selects an exploration goal as a function of the semantic map and the goal object with a learned exploration policy, and plans low-level actions to reach this goal. We replicate the exploration policy architecture and training process of \cite{chaplot2020object} for all datasets. We use Mask-RCNN \cite{chaplot2020object} pre-trained on MS-COCO for object detection and instance segmentation. The semantic map has a shape $K$ x $M$ x $M$ matrix where $M$ x $M$ $=$ $960$ x $960$ is the map size, with each cell corresponding to $25$ cm$^2$ ($5$ cm x $5$ cm) in the physical world, and $K = 16$ is the number of map channels. Semantic map features are computed with a convolutional neural network and passed through a feed-forward neural network along with a learnable embedding for the goal object to compute an exploration goal in $[0, 1]^2$, which is then converted to top-down map space. The exploration policy is trained for $10$ million steps using reinforcement learning with the Proximal Policy Optimization algorithm \cite{schulman2017proximal}, and the distance reduced to the nearest goal object as the reward. As in \cite{chaplot2020object}, we sample the long-term goal at a coarse time scale once every 25 steps.

\textbf{Scene Dataset Comparison.} \Cref{tab:objectnav_datasets} (bottom three rows) shows the performance of agents with a semantic exploration policy trained on \hmtdsem, Gibson, or MP3D training scenes and evaluated on each dataset's validation scenes. Agents trained on \hmtdsem scenes achieve the best validation performance averaged across all datasets. %Training on \hmtdsem scenes leads to the best-performing agents across all datasets.

% \setlength{\tabcolsep}{10pt}
% \begin{table*}
% \begin{minipage}{\linewidth}
%     \centering
%     {%
%     \begin{tabular}{@{}l|cc|cc|cc@{}}
%     \toprule
%     & \multicolumn{2}{c|}{MP3D (val)} & \multicolumn{2}{c|}{Gibson (val)} & \multicolumn{2}{c }{HM3D (val)} \\
%      Dataset                 & Success $\uparrow$ &  SPL $\uparrow$  &Success $\uparrow$ & SPL $\uparrow$   & Success $\uparrow$ & SPL $\uparrow$    \\ \hline
%     MP3D                    & 50.8 & 21.7 & 53.2 & 28.5 & $\mathbf{55.6}$ & $\mathbf{30.7}$ \\
%     Gibson-Medium           & 31.9 & 14.8 & $\mathbf{38.3}$ & $\mathbf{16.0}$ & $\mathbf{38.1}$ & $\mathbf{15.7}$ \\
%     HM3D                    & 45.3 & 20.5 & 47.1 & 21.5 & $\mathbf{49.4}$ & $\mathbf{23.4}$ \\ \bottomrule
%     \end{tabular}
%     }
%     \vspace{-5pt}
%     \caption{\textbf{Modular Learning Results.} ObjectNav performance of the modular semantic exploration policy when trained on on HM3D, Gibson, and MP3D and evaluated on all datasets.}
%     \label{tab:modular_learning_results} 
% \end{minipage}
% \end{table*}

%% file: sections/experiments-main-table.tex
\setlength{\tabcolsep}{15pt}
\begin{table*}
\begin{minipage}{\linewidth}
    \centering
    \resizebox{0.95\textwidth}{!}{%
    \begin{tabular}{@{}cl|cc|cc|cc|cc@{}}
    \toprule
    & Eval Dataset $\rightarrow$                        & \multicolumn{2}{c|}{Gibson (val)} & \multicolumn{2}{c|}{MP3D (val)} & \multicolumn{2}{c }{\hmtdsem (val)}  & \multicolumn{2}{c }{Average}\\
    Agent & Train Dataset $\downarrow$                 & Success $\uparrow$ &  SPL $\uparrow$  &Success $\uparrow$ & SPL $\uparrow$   & Success $\uparrow$ & SPL $\uparrow$ & Success $\uparrow$ & SPL $\uparrow$    \\ \midrule
    % \parbox[t]{2mm}{\multirow{3}{*}{\rotatebox[origin=c]{90}{RL}}}
    \multirow{3}{*}{RL}
    % & Gibson-Tiny             & 24.35 & 9.27 & 10.64 & 4.32 & 18.22 & 6.24 \\
    & Gibson           & 25.41 & 10.22 & \textbf{18.18} & 6.51 & 28.79 & 11.32 & 24.13 & 9.35 \\
    & MP3D                    & 19.66 & 7.16 & 17.53 & 6.59 & 18.97 & 6.93 & 18.70 & 6.89 \\
    & \hmtdsem                & \textbf{32.96} & \textbf{13.56} & 17.53 & \textbf{7.48} & \textbf{39.36} & \textbf{18.02} & $\mathbf{29.95}$ & $\mathbf{13.03}$ \\
    \midrule
    % IL
    % \parbox[t]{2mm}{\multirow{3}{*}{\rotatebox[origin=c]{90}{IL}}}
    \multirow{3}{*}{IL}
    & Gibson             & 18.15 & 7.84 & 16.57 & 5.84 & 20.06 & 6.03 & 18.26 & 6.57\\
    & MP3D                    & 18.95 & 6.95 & $\mathbf{28.57}$ & $\mathbf{9.91}$ & 25.37 & 7.73 & 24.30 & 8.20\\
    & \hmtdsem                &$\mathbf{28.02}$ &$\mathbf{11.64}$ & $25.06$ & $8.72$ &$\mathbf{33.49}$ &$\mathbf{11.64}$ & $\mathbf{28.86}$ & $\mathbf{10.66}$\\
    \midrule
    % ML
    % \parbox[t]{2mm}{\multirow{3}{*}{\rotatebox[origin=c]{90}{ML}}}
    \multirow{3}{*}{ML}
    & Gibson           & $\mathbf{38.3}$ & $\mathbf{16.0}$ & 47.1 & 21.5 & 53.2 & 28.5 & 46.2 & 22.0 \\
    & MP3D                    & 31.9 & 14.8 & 45.3 & 20.5 & 50.8 & 21.7 & 42.67 & 19.0 \\
    & \hmtdsem                & 38.1 & 15.7 & $\mathbf{49.4}$ & $\mathbf{23.4}$ & $\mathbf{55.6}$ & $\mathbf{30.7}$ & $\mathbf{47.70}$ & $\mathbf{23.27}$ \\
    \bottomrule
    \end{tabular}
    }
    \vspace{-5pt}
    \caption{\small \textbf{ObjectNav generalization across datasets with different agents.} We report results for a DD-PPO policy (RL), a Behavior Cloning policy (IL), and a modular semantic exploration policy (ML).\vspace{-10pt}}
    \label{tab:objectnav_datasets}
\end{minipage}
\end{table*}

%% file: sections/challenge.tex
\subsection{ObjectNav Challenge 2022}
The Habitat ObjectNav challenge~\cite{habitatchallenge2022} in the Embodied AI workshop in CVPR 2022 used the \hmtdsem dataset. The challenge received a total of 1022 submissions from 54 teams through the course of the challenge. This is much higher in comparison to the 2021 and 2020 Habitat ObjectNav challenge which received a total of 400 and 563 submissions from 45 and 27 teams respectively. The task definition was identical between the 2020-21 and 2022 challenges, and the only change was the dataset from Matterport3D~\cite{chang2017matterport3d} to HM3DSEM. The increase in participation highlights the importance of improving dataset scale and quality for community adoption. We report the final Success Rate and SPL of the submission from the top 5 teams and the DDPPO baseline on the Test-Challenge split in \Cref{tab:objectnav_challenge_results}.

% In the beginning of 2022, we released the Habitat ObjectNav challenge based on our newly released \hmtdsem dataset. We follow the procedure from \cite{batra2020objectnav} and released three phases of the challenge including Minival, Test Standard and Test Challenge. This year, we saw participation from 54 teams, receiving a total of 1022 submissions through the course of the challenge. This is much higher in comparison to the 2021 and 2020 ObjectNav challenge, where we had seen a total of 400 and 563 submissions from 45 and 27 teams respectively. Between the 2020-21 and 2022 challenge, the task definition was identical and only change was the dataset from Matterport3D~\cite{chang2017matterport3d} to \hmtdsem. The increase in participation highlights the importance of improving dataset scale and quality in community adoption. We present the final Success Rate and SPL of the submission from the top 5 teams and the DDPPO baseline on the Test-Challenge split in \Cref{tab:objectnav_challenge_results}.
%  The aim of the challenge is to benchmark the progress in the field of embodied AI.
\setlength{\tabcolsep}{8pt}
\renewcommand{\arraystretch}{1.05}
\begin{table}
\begin{minipage}{\linewidth}
    \centering
    \resizebox{\linewidth}{!}{%
    \begin{tabular}{@{}clccccc@{}}
    \toprule
        &\textbf{Team Name} && \textbf{Success Rate} (\%) $\uparrow$ && \textbf{SPL} (\%) $\uparrow$ &\\
        \midrule
        &ByteBOT && 64.0 && 35.0 &\\
        &BadSeed && 65.0 && 33.0 &\\
        &elf        && 61.0 && 30.0 &\\ 
        &Populus A. && 60.0 && 30.0 &\\
        &Stretch    && 56.0 && 29.0 &\\
        \midrule
        &DDPPO && 25.0 && 12.0 &\\
        \bottomrule
    \end{tabular}
    }
    \vspace{-5pt}
    \caption{\small \textbf{Habitat 2022 ObjectNav Challenge.} Performance of the top entries to the Habitat 2022 ObjectNav Challenge on the Test Challenge split. Entries are ordered by SPL. 
    }
    \label{tab:objectnav_challenge_results} 
    \vspace{-15pt}
\end{minipage}
\end{table}\textbf{}

%% file: sections/conclusion.tex
\vspace{-10pt}
\section{Conclusion}

We present the \hmtdsem dataset which is the largest public dataset of real-world spaces with dense semantic annotations. Unlike prior datasets, \hmtdsem uses texture information to annotate pixel-accurate object boundaries. The dataset has undergone an intense expert annotation as well as a verification process to maximize accuracy and coverage. All scene annotations are provided in a standardized format, making it easy to use with the existing Habitat simulator. We demonstrate the effectiveness of the \hmtdsem dataset for the Object Goal Navigation task using reinforcement learning, imitation learning, and modular learning methods. Across different methods, we show the importance of improved annotation quality and larger datasets by showing that policies trained using \hmtdsem outperform the policies trained on prior datasets and the performance of policies improves as we increase the training dataset size. The introduction of the \hmtdsem in the Habitat ObjectNav challenge in 2022 led to a significant increase in participation. We hope that the high quality and scale of HM3DSem spurs future progress in Embodied AI. 

%We present the Habitat-Matterport 3D Semantics (HM3DSEM) dataset.
%HM3DSEM is the largest dataset of real-world spaces with densely annotated semantics that is currently available to the academic community. The scale, quality, and diversity of object annotations far exceed those of datasets from prior work. A key difference setting apart HM3DSEM from other datasets is the use of texture information to annotate pixel-accurate object boundaries

%% file: sections/acknowledgments.tex
\section*{Acknowledgements}
The Georgia Tech effort was supported in part by NSF, ONR YIP, and ARO PECASE. The views and conclusions contained herein are those of the authors and should not be interpreted as necessarily representing the official policies or endorsements, either expressed or implied, of the U.S. Government, or any sponsor.

%% file: sections/appendix.tex
\section{Appendix Annotation Inferences}
\subsection{Assumptions}
To derive accurate and reasonable inferences for the HM3D dataset annotations, we make the following assumptions about annotations and layouts:
\begin{itemize}
    \item Objects: Annotations accurately describe the object being annotated.
    \item Objects: Object annotations with qualifiers in their name, such as “bath xxx” or “kitchen xxx” infer strongly where these objects are found.(i.e. bathrooms or kitchens, respectively)
    \item Objects: Annotated objects are generally not staged to be “unnatural” in their configuration but rather the object layouts are natural reflections of human use (i.e. we would not expect to see the same region holding a toilet, a refrigerator, a stovetop and a bed).
    \item Regions: Region annotations are derived from reasonable estimates of room boundaries within the scenes, (i.e. two objects with the same region annotation could be said to be “in the same room”)
    \item Scenes:  Scenes are generally reasonable representations of actual human environments and are not staged in some unnatural way (scene full of bathrooms, for instance). We do allow for non-habitation scenes, such as office spaces or restaurants.
\end{itemize}

Using these assumptions we analyze the semantic annotation text files to learn about the nature of the underlying dataset.  We can infer relationships between objects based on mutual regional membership, properties of the regions that contain these objects, and even gain some insight into the nature of the scenes themselves.

\subsection{Instance Segmentation and Object Detection}

\begin{table}[H]
    \centering
    % \hspace*{-0.1in}
    \resizebox{\linewidth}{!}{
        \begin{tabular}{@{}lP{0.8cm}P{0.8cm}P{0.8cm}P{1.1cm}P{0.8cm}P{0.8cm}P{0.8cm}P{1.1cm}@{}}
            \toprule
                            & \multicolumn{4}{c}{Object detection (mAP@0.5)} & \multicolumn{4}{c@{}}{Instance segmentation (mAP@0.5)} \\ \cmidrule(lr){2-5} \cmidrule(lr){6-9}
 Training $\downarrow$      & Gibson &  MP3D   &  HM3D   &  ADE20k & Gibson  &   MP3D  &  HM3D   &  ADE20k \\ \midrule
 Gibson                     &  24.4  &  2.1    &  3.1    &  1.1       &  23.5   &  0.7    &  1.3    &   0.6   \\
 MP3D                       &  26.9   & 27.3   &  31.7   &  14.5      &  21.5   & 20.1    &  25.9   &  10.0      \\
 \hmtdsem                   &\tb{39.6}&\tb{33.7}&\tb{54.0}&\tb{31.8}   &\tb{35.2}&\tb{28.2}&\tb{50.1}&\tb{26.4}\\

            \bottomrule
            \end{tabular}
    }
    \vspace{-0.1in}
    \caption{\small \textbf{Benchmarking object detection, instance segmentation:} We learn Mask-RCNN models on each train dataset (column 1) and evaluate them on all val datasets (columns 2-9). We also test on real-world images from ADE20k. Training on \hmtdsem leads to the best generalization to new scenes and datasets. }
    \label{tab:segmentation}
    \vspace{-0.1in}
\end{table}

\begin{figure}[t]
    \centering
    \hspace*{-0.1in}
    \includegraphics[width=0.52\textwidth,clip,trim={0 8.1cm 3.5cm 0}]{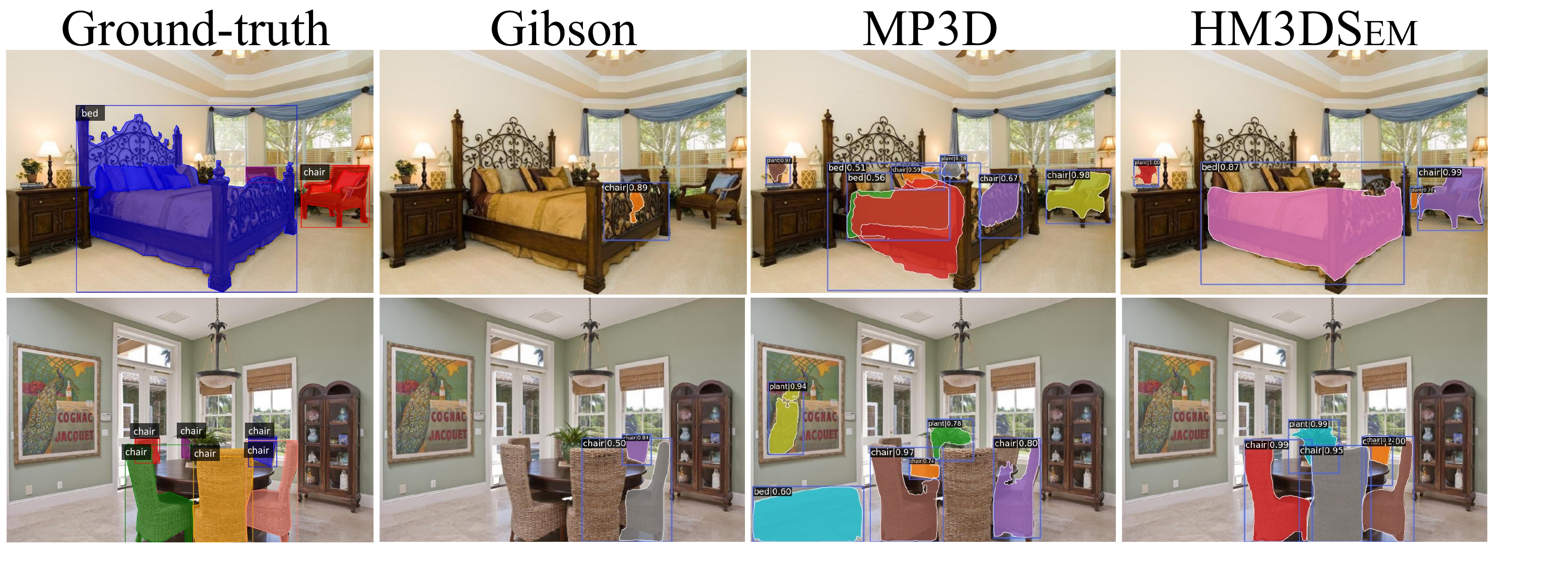}
    \vspace*{-0.3in}
    \caption{\small \textbf{Visualizing instance segmentation.} In each row, we display an image sampled from ADE20k with overlayed masks denoting the ground-truth (column 1) and model predictions (columns 2-4). The \hmtdsem-trained model generalizes to ADE20k much better than those trained on Gibson and MP3D.}
    \vspace*{-0.1in}
    \label{fig:qual}
\end{figure}

We rendered instance segmentation annotations for 150k train, 10k val images from each of \hmtdsem, Gibson, and MP3D. For each dataset, we train a Mask-RCNN for 10 epochs to predict the 6 object classes used in ObjectNav:  chair, bed, plant, toilet, tv/monitor, and sofa (similar to~\cite{chaplot2020object}). We then evaluate each model on all three val splits for object detection and instance segmentation. We additionally test on $\sim$500 real-world images of residential scenes from the ADE20k dataset. See~\cref{tab:segmentation}. The model trained on \hmtdsem generalizes best across scenes and datasets by a large margin. This echoes our ObjectNav results and reaffirms the value of \hmtdsem for visual perception. Training on Gibson leads to poorest performance due to annotation inaccuracies and sparsity. We visualize examples in~\cref{fig:qual}.

\subsection{Analysis Method}
To accomplish this analysis, we hand assigned a region proposal to 261 of the 1624 unique annotation tags provided by the annotators heuristically based on the annotation name. These region annotation proposals are not treated as absolutes, but rather suggestions - if an object found within some region’s category tag is mapped to a specific proposal, this proposal serves only to suggest that the containing region might be described using this annotation. In this way, instead of expecting a direct labeling, the objects’ region proposals are used as votes for their containing regions’ possible annotations. 

The possible region names chosen for this experiment (and the category tags mapped to each) are listed below: 
\begin{itemize}
    \item \textbf{bathroom}: bath,  bath bar,  bath cabinet,  bath carpet,  bath cosmetics,  bath curtain,  bath curtain bar,  bath dial,  bath door,  bath door frame,  bath faucet,  bath floor,  bath grab bar,  bath hanger,  bath mat,  bath shelf,  bath shower,  bath side table,  bath sink,  bath tap,  bath towel,  bath towels,  bath tub,  bath utensil,  bath wall,  bathmat,  bathrobe,  bathroom accessory,  bathroom art,  bathroom cabinet,  bathroom cabinet door,  bathroom cabinet drawer,  bathroom counter,  bathroom floor,  bathroom glass,  bathroom mat,  bathroom rug,  bathroom shelf,  bathroom stuff,  bathroom towel,  bathroom utensil,  bathroom utensils,  bathroom wall,  bathroom window,  bathtub,  bathtub knob, bathtub platform, bathtub tap, bathtub utensil, bidet, shower, shower bar, shower base, shower battery, shower bench, shower cabin, shower cabinet, shower caddy, shower case, shower ceiling, shower ceiling lamp, shower cockpit, shower cosmetics, shower curtain, shower curtain bar, shower curtain rod, shower dial, shower door, shower door frame, shower door knob, shower floor, shower frame, shower glass, shower grab bar, shower handle, shower handrail, shower hanger, shower hose, shower hose/head, shower knob, shower mat, shower mirror, shower pipe, shower rail, shower rod, shower seat, shower shelf, shower soap shelf, shower stall, shower step, shower tap, shower tray, shower tub, shower utensil, shower valve, shower wall, shower wall cubby, shower window frame, shower-bath cabinet, showerhead, toilet, toilet brush, toilet brush holder, toilet cleaner, toilet paper, toilet paper dispenser, toilet seat, toothbrush, toothpaste, wall toilet paper
    \item \textbf{bedroom}: bed, bed base, bed cabinet, bed cabinet lamp, bed comforter, bed curtain, bed ladder, bed light, bed sheet, bed small, bed stand, bed table, bedding, bedframe, bedpost, bedroom ceiling, bedroom table, bedside cabinet, bedside cabinet door, bedside cabinet drawer, bedside lamp, bedside table, ceiling bedroom, dresser, jewelry box, nightstand, wardrobe
    \item \textbf{dining room}: dining chair, dining table, dinner chair, dinner table
    \item \textbf{garage}: garage door, garage door frame, garage door motor, garage door opener, garage door opener bar, garage door opener motor, garage door opener railing, garage door railing, garage light
    \item \textbf{hall/stairwell}: stair, stair frame, stair handle, stair step, stair wall, staircase, staircase handrail, staircase trim, staircase wall, stairs, stairs railing, stairs skirt, stairs trim, stairs wall, stairwell
    \item \textbf{kitchen}: cabinet kitchen, dish rack, dishwasher, fridge, kitchen appliance, kitchen cabinet, kitchen cabinet door, kitchen cabinet drawer, kitchen cabinet lower, kitchen ceiling, kitchen chair, kitchen counter, kitchen counter item, kitchen counter support, kitchen countertop item, kitchen countertop items, kitchen decoration, kitchen extractor, kitchen gloves, kitchen handle, kitchen island, kitchen knife set, kitchen lower cabinet, kitchen lower shelf, kitchen seating, kitchen shelf, kitchen sink, kitchen sink cabinet, kitchen table, kitchen top, kitchen towel, kitchen utensil, kitchen utensils, kitchen wall, kitchenware, knife holder, knife set, oven, oven and stove, refrigerator, refrigerator cabinet, stove, stovetop
    \item \textbf{laundry room}: washer-dryer, washing machine, washing machine and dryer, washing powder, washing stuff
    \item living room: circular sofa, coffee table, couch, l-shaped sofa, recliner, remote control, sofa, sofa chair, sofa seat, sofa set
    \item \textbf{office}: computer, computer chair, computer desk, computer equipment, computer mouse, computer tower, desk, desk cabinet, desk chair, desk clutter, desk door, desk lamp, desk organizer, laptop, office chair, office table, office wall
    \item \textbf{rec room}: barbell, exercise ball, exercise bike, exercise equipment, exercise ladder, exercise machine, exercise mat, exercise mat roll, exercising blocks, foosball game table, foosball table, gym equipment, gym mat, gym rope, gym stepper, pool stick, pool table, rack of weights, weight bench, yoga mat
\end{itemize}

We then record the category tag for every object instance in the dataset, along with region annotation proposals for all categories that have been assigned them, on a per tag (not per object instance) basis, and organize this data per region per scene. 

\subsection{Scene-level Statistics}
Some observations about all 216 scenes were made using the region labeling heuristics and examining category presence as well as room annotations derived from proposal votes. 

Using only proposal-tagged category presence as a guide, we found:
\begin{itemize}
    \item 12 scenes lacked any objects containing tags with the “bedroom” proposal. These scenes were all visually verified to be commercial spaces, either offices, restaurants, or stores. 
    \item 7 scenes lacked category tags with the “bathroom” proposal. These were also visually verified to be non-residential spaces. 1 scene had instances of “bedroom” categories but none of “bathroom”; this scene is a large house that has been converted to a museum.
    \item 25 scenes contained objects with proposed “garage” region annotations. These were visually verified to all contain garages.
    \item 9 scenes lacked any objects with proposed “kitchen” region annotations. 4 of these were commercial spaces that also lacked “bedroom” proposals, while 4 of the remaining 5 were hotel rooms or suites. The last remaining scene was found to actually contain a kitchen through visual inspection, which had a modern design, lacking most obvious appliances such as “stove” or “oven”; however, a refrigerator was present and visible but was mislabeled as a “cabinet”.
\end{itemize}

By aggregating votes per region of the number of category-derived region proposals, we derived potential room labels, which provided even more accurate suggestions of scene content, as shown in section 3.4.

\subsection{Region Label Inference}

\begin{figure*}[ht]
    \centering
    \includegraphics[width=0.95\textwidth]{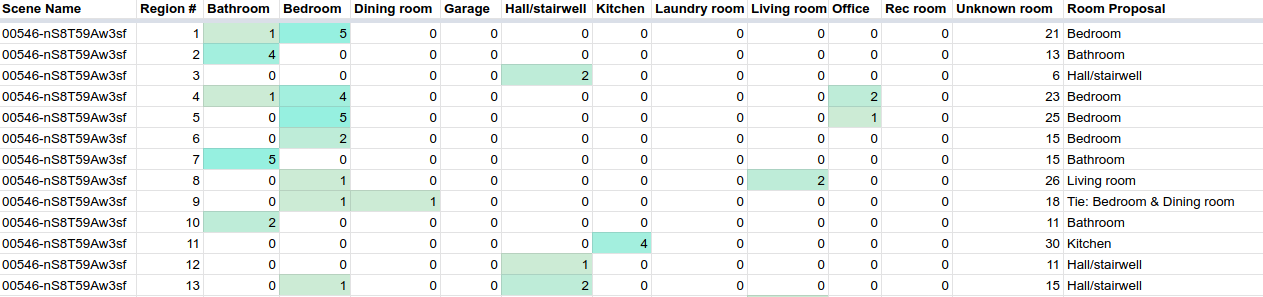}
    \caption{Scene 00546-nS8T59Aw3sf Region label proposals based on category presence}
    \label{fig:region_votes_00546}
    \vspace{-6pt}
\end{figure*}
It would be useful if the region proposal votes derived from each region’s constituent object categories could be used to yield labels for the region itself. By randomly picking scenes, the legitimacy of this data for region labeling could be investigated through visual verification in the Habitat engine. \Cref{fig:region_votes_00546} shows the aggregation results for a randomly chosen scene.

Using the highest vote counts per region/row to suggest that region’s proposed annotation, this scene’s 13 regions are proposed to be 3 bathrooms, 4 bedrooms, 3 hallway/stairwells, 1 kitchen, 1 living room, 1 of either bedroom or dining room. Visually inspecting this room yields the same count, with the confused room being the dining room. A serving buffet is mislabeled in this room as a dresser. 

\begin{figure*}[ht]
    \centering
    \includegraphics[width=0.95\textwidth]{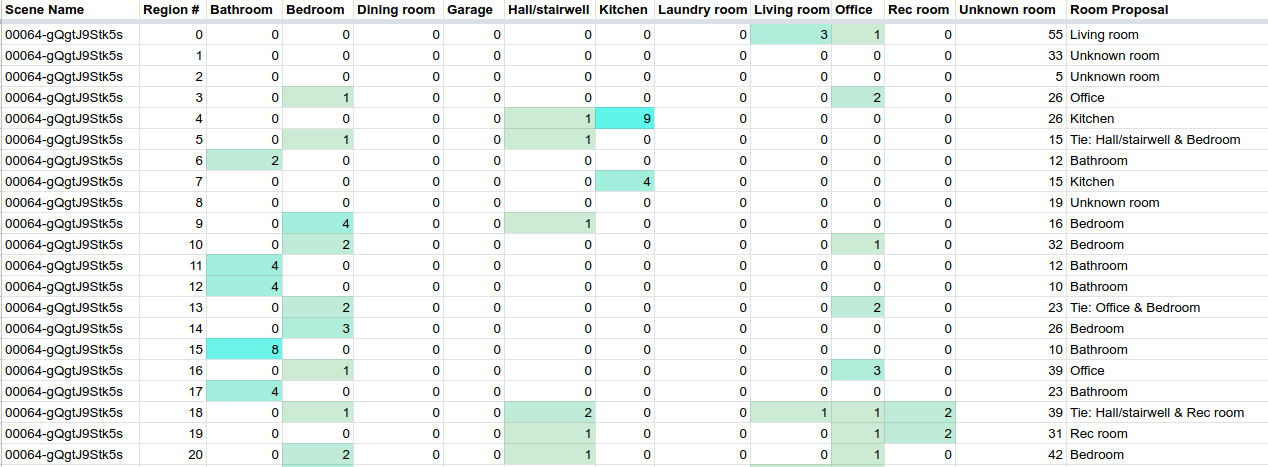}
    \caption{Scene 00064-gQgtJ9Stk5s Region label proposals based on category presence}
    \label{fig:region_votes_00064}
    \vspace{-6pt}
\end{figure*}

Even for larger scenes with many regions, the per-region proposal aggregations can provide useful insights. \Cref{fig:region_votes_00064} shows the results for a larger scene.

Note that 3 of the 21 regions in this scene lack specific proposal votes (regions 1,2,8) due to the categories of the objects found in these regions not having region proposals. Of the 18 regions with proposals, 5 bathrooms (6,11,12,15,17), 4 bedrooms(9,10,14,20), 1 living room(0), 2 offices(3 ,16), 1 rec room(19) and 2 kitchens(4,7), along with 3 ambiguous mappings of 1 bedroom or office (13), 1 hallway or bedroom (5), and 1 hallway or rec room (18). 

Visually inspecting this scene yields very similar results:   2 rec rooms(regions 18,19), 2 kitchen(4,7), 1 dining room(1), 1 office(3), 3 stairs/hallways (2,5,8), 5 bathrooms (6, 11,12, 15,17), 6 bedrooms (9,10, 13, 14, 16, 20).

\begin{figure*}[ht]
    \centering
    \includegraphics[width=0.95\textwidth]{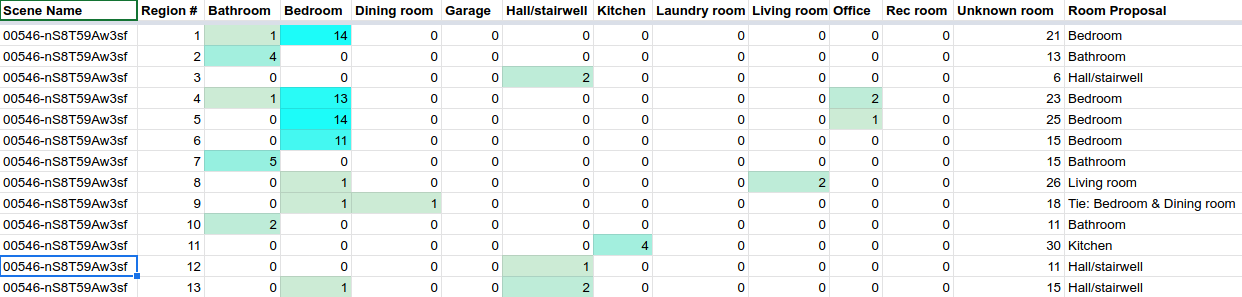}
    \caption{Scene 00546-nS8T59Aw3sf Region label proposals based on category presence with weighting}
    \label{fig:region_weighted_votes_00546}
    \vspace{-6pt}
\end{figure*}

\begin{figure*}[ht]
    \centering
    \includegraphics[width=0.95\textwidth]{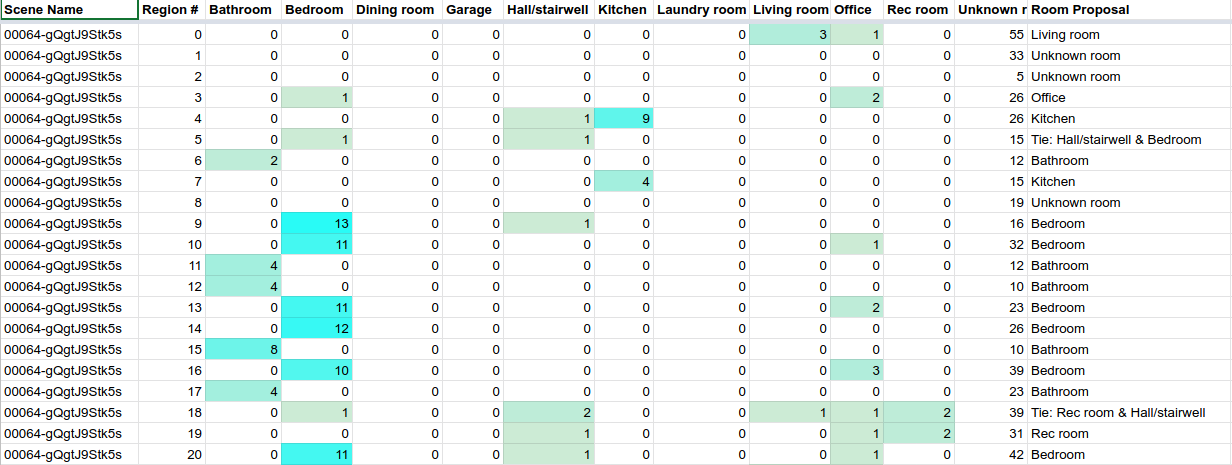}
    \caption{Scene 00064-gQgtJ9Stk5s Region label proposals based on category presence with weighting}
    \label{fig:region_weighted_votes_00064}
    \vspace{-6pt}
\end{figure*}

Ambiguity in the proposal assignments can be mitigated if certain categories, such as “bed”, received more votes, although this might miscategorize regions where beds were in storage. \Cref{fig:region_weighted_votes_00546} and \Cref{fig:region_weighted_votes_00064} show these same two scenes with "bed" category receiving 10 votes for "bedroom" proposal instead of 1.

Using these region proposals at the scene level gives a reasonable estimate for the room layout and count of each scene. 
% \todo:  scene layout/room proposal graphic

\subsection{Object-Level Analysis and Files}

Along with collecting and organizing object instance-based category data organized by scene and then region, we also aggregated the scene, region and per-region neighbor categories for each category present across the entire dataset, so that the categories of all objects that share a region are known to one another, as are the region annotation proposals for those categories that have them.

The statistics on category prevalence throughout all the regions in the dataset provides suggestions for possible region proposal labels for otherwise unmapped category tags based on the category “company they keep”.

We have provided the following files to assist users in conducting their own analyses of the semantic scenes. 
\begin{itemize}
    \item \textbf{HM3D\_CountsOfObjectTypes.csv} : This file provides the name and number of occurrences of every category label in use across the entire dataset.
    \item \textbf{Per\_Category\_Counts\_Uncommon.csv} : This file provides the number of occurrences of every category label not including common architectural components (excluding doors/walls/ceilings/etc)
    \item \textbf{Per\_Scene\_Neighborhood\_Stats.csv} : This file contains scene-level category statistics (mean, variance, skew, kurtosis) describing number of regions per scene and the unique categories and object instances they contain.
    \item \textbf{Per\_Scene\_Region\_Neighborhoods.csv} : This file contains per-scene-per-region unique category and instance counts, and all object instance labels (including common labels) within each region.
    \item \textbf{Per\_Scene\_Region\_Votes.csv} : This file lists the per-scene-per-region votes for region/room label proposal based on the categories of the various object instances present with hand-annotated labels.  Each object instance of a category with a region proposal gets 1 vote.
    \item \textbf{Per\_Scene\_Total\_Votes.csv} : This file has the per-scene room label proposal counts (i.e. how many proposed bedrooms, bathrooms, etc. are present in a scene), built from the “Per\_Scene\_Region\_Votes.csv” data.
    \item \textbf{Per\_Scene\_Region\_Weighted\_Votes.csv} : This file also lists the per-scene-per-region votes for region/room label proposal based on the categories of the various object instances present with hand-annotated labels, except in this case, instances of the category “bed” receive 10 votes.  All other object instances of categories with assigned region proposals still receive 1 vote.
    \item \textbf{Per\_Scene\_Total\_Weighted\_Votes.csv} : This file has the per-scene room label proposal counts (i.e. how many proposed bedrooms, bathrooms, etc. are present in a scene), built from the “Per\_Scene\_Region\_Weighted\_Votes.csv” data, where instances of the “bed” category received 10 votes instead of just 1.
    \item \textbf{Region\_Tag\_Mappings.csv} : This file lists per-scene-per-region count of categories present, and names of "uncommon" tags (excluding common architectural categories like wall, ceiling/etc)
\end{itemize}

The following files include the region proposal aggregate categories in their reporting. Each of these aggregate categories are formed by a union of all the categories that share the same hand-annotated region proposal.

\begin{itemize}
    \item \textbf{Per\_Category\_Region\_Neighbors.csv} : This file provides statistics for category and instance presence in scenes and regions.  This includes the number of scenes and number of regions that instances of the category are present, as well as the number of instances total and the average number of instances per scene and per region when present.  The total number of unique neighbor categories, where a neighbor is defined as sharing the same region, for each category is also listed as well as the categories and region counts of each neighbor.

    \item \textbf{Per\_Category\_Region\_Per\_Cat\_Votes.csv} : This file lists the per-category hand-assigned region proposal tags (bed inferring bedroom, for example), if present, as well as the counts of other neighbor categories’ hand-labeled region proposals. This is useful in determining the types of regions where instances of categories are most likely to be found. For example, the “air vent” category shares regions with 206 instances of “bathroom”-labeled categories, 49 instances of “bedroom”-labeled categories, 61 instances of “kitchen”-labeled categories, etc. 

    \item \textbf{Per\_Scene\_Region\_Cat\_Presence.csv} : This file holds per-scene-per-region unique category presence and count of instances of each category.
\end{itemize}